\documentclass[journal]{IEEEtran}

\ifCLASSINFOpdf

\else

\fi

\usepackage[belowskip=-15pt,aboveskip=0pt]{caption}
\usepackage{amssymb}
\usepackage{algorithmic}
\usepackage{float}
\usepackage{lipsum}
\usepackage{ifmtarg}
\usepackage{array}
\usepackage{amsmath}

\usepackage{color}
\usepackage{tabularx}
\usepackage[colorlinks=true, allcolors=green]{hyperref}
\usepackage{adjustbox}
\usepackage{epstopdf}
\usepackage{caption}
\usepackage[ruled,linesnumbered,vlined,commentsnumbered]{algorithm2e}
\usepackage{cite}
\usepackage{booktabs}
\usepackage{multirow}

\usepackage{caption}
\usepackage{subcaption}
\graphicspath{{figures/}}

\def\hl{\setlength{\fboxsep}{1.0pt}\colorbox[rgb]{0.85,0.85,0.85}}

\begin{document}

\title{An Ensemble Semi-Supervised Adaptive Resonance Theory Model with Explanation Capability for Pattern Classification}

\author{Farhad~Pourpanah,~\IEEEmembership{Member,~IEEE,}
        Chee~Peng~Lim,
        Ali~Etemad,~\IEEEmembership{Senior Member,~IEEE},
        and Q.~M.~Jonathan~Wu,~\IEEEmembership{Senior Member,~IEEE}
        
\thanks{This work is partially funded by the NSERC CREATE  Program on TrustCAV (\textit{Corresponding author: Q. M. Jonathan Wu}).}
\thanks{F. Pourpanah and A. Etemad are with the Department of Electrical and Computer Engineering, Queen’s University, Kingston, ON, Canada (e-mail: f.pourpanahnavan@queensu.ca \& ali.etemad@queensu.ca).}
\thanks{C. P. Lim is with Institute for Intelligent Systems Research and Innovation, Deakin University, Australia (e-mail: chee.lim@deakin.edu.au).}
\thanks{Q. M. Jonathan Wu is with the Department of Electrical and Computer Engineering, University of Windsor, Windsor, ON N9B 3P4, Canada (e-mail: jwu@uwindsor.ca).}
}

\maketitle

\begin{abstract}
Most semi-supervised learning (SSL) models entail complex structures and iterative training processes as well as face difficulties in interpreting their predictions to users. To address these issues, this paper proposes a new interpretable SSL model using the supervised and unsupervised Adaptive Resonance Theory (ART) family of networks, which is denoted as SSL-ART. Firstly, SSL-ART adopts an unsupervised fuzzy ART network to create a number of prototype nodes using unlabeled samples. Then, it leverages a supervised fuzzy ARTMAP structure to map the established prototype nodes to the target classes using labeled samples. Specifically, a one-to-many (OtM) mapping scheme is devised to associate a prototype node with more than one class label. The main advantages of SSL-ART include the capability of: \textit{(i)} performing online learning, \textit{(ii)} reducing the number of redundant prototype nodes through the OtM mapping scheme and minimizing the effects of noisy samples, and \textit{(iii)} providing an explanation facility for users to interpret the predicted outcomes. In addition, a weighted voting strategy is introduced to form an ensemble SSL-ART model, which is denoted as WESSL-ART. Every ensemble member, i.e., SSL-ART, assigns {\color{black}a different weight} to each class based on its performance pertaining to the corresponding class. The aim is to mitigate the effects of training data sequences on all SSL-ART members and improve the overall performance of WESSL-ART. The experimental results on eighteen benchmark data sets, three artificially generated data sets, and a real-world case study indicate the benefits of the proposed SSL-ART and WESSL-ART models  for tackling pattern classification problems.

\end{abstract}

\begin{IEEEkeywords}
Adaptive resonance theory, semi-supervised learning, incremental learning, weighted voting ensemble, rule extraction, anomaly detection, data classification 
\end{IEEEkeywords}

\IEEEpeerreviewmaketitle

\section{Introduction}
\label{sec:Int}
\IEEEPARstart{S}{upervised} learning models are effective for problems with sufficient labeled samples.
However, obtaining labeled samples in many real-world problems, e.g., medical diagnosis and fault diagnosis, is expensive, difficult, and time-consuming, as it requires human labeling efforts.
Meanwhile, unlabeled samples contain useful information such as geometric structure, and exploiting such information can help to learn a better model~\cite{wang2020recent,pourpanah2022review,abdar2021review,orouskhani2022alzheimer}.
Semi-supervised learning (SSL) techniques offer a good compromise to address such problems~\cite{gu2020asemi,pourpanah2021asemisupervised,jia2020semisupervised}.
They use a large number of unlabeled samples to augment a small set of labeled samples, in order to improve the performance of a learning model~\cite{Silva2012Network,jing2020group,zhong2020semisupervised}.

The success of SSL techniques is attributed to the three assumptions, which are semi-supervised smoothness, manifold, and cluster assumptions~\cite{Chapelle2010Semi}.
Semi-supervised smoothness assumes that two samples should share the same target class if they are close in a high-density region.
The manifold assumption states that a higher degree of similarity is ensured if the high-dimensional samples lie on a low-dimensional manifold.
While the cluster-based assumption indicates that if two samples are located in a cluster, they belong to the same class~\cite{Chen2011Semi}.
In general, SSL techniques can be categorized into self-training, co-training, generative and graph-based methods~\cite{Zhu2005Semi,van2020survey}. 
Self-training techniques use a classifier to bootstrap the learning model with labeled samples that obtained from their own highly confident predictions~\cite{rosenberg2005semi}.
While, co-training techniques utilize multiple classifiers and iteratively re-train them by each other’s most confident predictions~\cite{abdelgayed2017fault}.
Generative-based methods aim to model a process to generate samples~\cite{xie2021semi}. 
Graph-based techniques build graph and find a minimum energy to link unlabeled samples to the labeled ones~\cite{ienco2020enhancing}.
Although the existing SSL methods have shown good performance on different problems, most of them behave as a black box due to their complex structures and iterative process, leading to difficulties for users to interpret their predictions.

The family of Adaptive Resonance Theory (ART) networks~\cite{Grossberg1976Adaptive} offers a self-organized, online learning model pertaining to a sequence of incoming data samples with the capability of providing explanation pertaining to a predicted output and solving both unsupervised and supervised problems.
Fuzzy ART~\cite{Carpenter1991Fuzzy} and fuzzy ARTMAP~\cite{Carpenter1992FuzzyARTMAP} are the most prominent members of unsupervised and supervised ART models, respectively.
They benefit from ART in solving stability-plasticity dilemma, i.e., the ability to learn new samples without forgetting previously learned ones. 
They also benefit from fuzzy logic in handling vague and imprecise human linguistic information.
The key learning concept of an ART model is the determination of the similarity level among the prototype nodes stored in its structure and the current input sample against a criterion.
If the criterion is not satisfied, the ART model adds a new prototype node to encode the current input sample.
To date, many hybrid models have been developed to enhance the performance of ART-based networks, e.g., fuzzy ARTMAP with reinforcement learning~\cite{Pourpanah2016Hybrid,pourpanah2019reinforced},
population-based evolutionary algorithms~\cite{Pourpanah2018non,pourpanah2017aqlearning}, 
as well as SSL method~\cite{Nooralishahi2018Semi}.
However, one key limitation of ART-based models is a high dependence on the sequence of the learning samples.
\vspace{-0.3cm}

\subsection{Contributions}

In this study, we propose a new ensemble-based SSL model with an explanatory capability based on the family of ART networks. The main contributions of our study are as follows:

\begin{itemize}
    \item {\color{black} A new two-stage SSL model, i.e. SSL-ART, which is able to perform incremental learning and add new prototype nodes dynamically into its structure when necessary, is introduced. Unlike most SSL models reviewed in survey papers~\cite{Zhu2005Semi,van2020survey} that use an iterative process to label the unlabeled samples before using all data samples for training purposes, SSL-ART requires only a one-pass learning procedure through both labeled and unlabeled samples.
    Besides requiring only one prototype (aka hidden) layer in its structure, SSL-ART is able to generate the required number of prototype (hidden) nodes according to a vigilance threshold to solve the underlying classification problem without pre-defining the network architecture. This capability of generating a dynamic network structure incrementally is important to avoid over-fitting or under-fitting issues caused by an inappropriate pre-defined number of hidden nodes/layers in most neural network-based learning models;}

    \item SSL-ART is equipped with a one-to-many (OtM) mapping scheme. 
    It enables the prototype nodes to establish an association with more than one target class. 
    As a result, the number of redundant prototype nodes is minimized, resulting in better performances in tackling noisy data problems;
    
    \item SSL-ART is able to produce a set of user-comprehensible If-Then rules in fuzzy linguistic variables for justification of its predictions.  
    As such, users can interpret the predicted outcomes in linguistic terms such as ``small", ``medium" and ``large" for reasoning purposes.
    This explanatory capability is critical in decision support tasks, particularly for safety-related applications such as medical and/or industrial diagnosis and prognosis problems;
    
    \item A weighted ensemble SSL-ART (WESSL-ART) model with a voting strategy is formulated.  
    Each ensemble member assigns a different weight to each class based on the associated performance metrics pertaining to the corresponding class.  
    WESSL-ART is a useful ensemble-based SSL model that is able to reduce the effects of the sequence of training samples during online learning, as indicated through a series of empirical studies.  
    
\end{itemize}

\subsection{Organization}

This paper contains five sections. {\color{black}Section~\ref{Sec:Related} provides a review on both online and semi-supervised learning models. Section~\ref{Sec:Pre} presents details of the fuzzy {\color{black}subsethood} measure and dynamics of fuzzy ART and fuzzy ARTMAP models. }
The proposed SSL-ART and ESSL-ART models are explained in Section~\ref{Sec:Method}.
The experimental study using benchmark and real-world problems is presented in Section~\ref{Sec:ExP}.
Concluding remarks are given in Section~\ref{sec:Conclusion}.

\vspace{-0.2cm}
\section{Related works}
\label{Sec:Related}
In this section, we provide a review on online (incremental) learning and semi-supervised learning models.

\subsection{Online (incremental) learning}
\label{Sec:online}

In general, learning in neural networks can be categorized into off-line (batch) and online (incremental) learning~\cite{jain2014review}. 
Online learning is susceptible to the problem of catastrophic forgetting~\cite{mccloskey1989catastrophic}, which is also known as the stability-plasticity problem.  
When a new input sample is presented, the model {\color{black} is required to} learn the sample in a one-pass manner, which can result in interference with previously learned network weights~\cite{pourpanah2019reinforced}.  
As such, circumventing the catastrophic forgetting problem is a key focus in online learning models.\par

ARCIKELM~\cite{tahir2020open} is an incremental learning model based on the extreme learning machine for food recognition. It dynamically adjusts the model structure in a way that reduces the catastrophic forgetting problem. 
In~\cite{guan2021reduce}, a self-supervised learning model with incremental learning capability was introduced.  
It addresses the dilemma of reducing the learning rate for retaining the existing knowledge. This model learns a task-specific self-supervised learning signal to help extracting features that are informative for both current and future tasks.  
A-iLearn~\cite{agarwal2022ilearn} is an ensemble-based incremental learning model for spoof fingerprint detection. It integrates an ensemble of base classification models trained on new data samples with the current ensemble model trained based on previous data samples.\par

In~\cite{yuan2022incremental}, the progressive fuzzy three-way concept is used to form an incremental learning model for recognizing objects in dynamic environments. Firstly, the fuzzy three-way concept is introduced by defining object and attribute learning operators. 
Then, the progressive process method is formulated to learn the corresponding concept for classifying objects. In [22], a model based on pseudo-rehearsal method for fault detection and classification subject to data streams with changes over time is proposed. 
For each fault type, a generative-rehearsal strategy is derived by combining a pseudo-rehearsal technique and independent generative models. 
IL-SSOR~\cite{chen2022incremental}, which is an incremental learning semi-supervised ordinal regression model, uses the Karush–Kuhn–Tucker (KKT) conditions to directly update the solutions of semi-supervised ordinal regression in tackling large-scale problems.\par

\vspace{-0.32cm}
\subsection{Semi-Supervised Learning}
\label{Sec:sec:SSL}
Recently, many SSL models have been developed. OT-OS~\cite{chau2021st_os} is an SSL model based on self-training for course-level prediction. 
It forms an ensemble model by combining self-training and tri-training to effectively learn from each selected set of unlabeled samples.  
In~\cite{nartey2019semi}, a selection mechanism was proposed.
It aims to prevent mistakes in reinforcement, which is a common problem in conventional self-training models. 
A co-training of two classification models was proposed in~\cite{abdelgayed2017fault} to detect faults in both transmission and distribution systems. 
Partial Bayesian co-training (PBCT)~\cite{nguyen2019partial} creates a partial view by scaling the original set of features, and then enhances the model performance by exploiting side information from partial data. 
CoTrade~\cite{he2020semisupervised} forms reliably communicate between two classifiers by a co-training algorithm. At each iteration, it considers the confidence of the classifier’s prediction and the number of predicted labels with high confidence of each classifier.\par

The proposed probabilistic SSL model in~\cite{fujino2008semisupervised} leverages the advantage of a generative model in its structure for solving single and multi-class labeled problems.  
The generative model is trained on labeled samples, while a biased correlation model is introduced.  
Both models use the same structure with different parameters. The two models are combined using the maximum entropy principle to form a hybrid classifier.  
In~\cite{chen2021robust}, a generative SSL model that is robust to outliers and noisy samples is proposed. This model, which is based on a variational autoencoder (VAE), improves its robustness by considering the uncertainty of the input samples. 
A denoising layer is also integrated into the VAE structure. BSOG~\cite{he2020semisupervised} is a graph-based SSL model that adaptively learns local manifold for bans selection. It optimizes the similarity matrix to build a reliable graph. 
A bipartite graph for SSL was introduced in~\cite{he2019fast}.  
It integrates anchors into the graph to capture the manifold structure and reduce the computation time. In addition, the bipartite graph structure is used to effectively deal with the out-of-distribution samples.\par

Although the reviewed methods have shown promising results, they suffer from several limitations.  
Self-training and co-training methods require an iterative {\color{black} process to label available unlabeled samples}, which consumes a long execution time.  
In addition, mislabeling any unlabeled samples can affect the generalization ability of supervised learning models. On the other hand, generative-based methods require the estimation of data distribution while graph-based methods normally result in a complex graph structure. 
Importantly, most of these studies are not able to perform incremental learning, comprising their effectiveness in real-world data streaming environments~\cite{pourpanah2021asemisupervised}.\par

Several studies have indicated that clustering unlabeled data can improve the performance of classification algorithms~\cite{dara2002clustering,goldberg2009multi}.  
As such, two-stage SSL models have been developed, in which an unsupervised learning model is used for pre-training while a supervised learning model is adopted for fine-tuning~\cite{pourpanah2021asemisupervised,hu2019utilizing}.  
In this study, we propose a two-stage SSL model that is able to incrementally learn from both labeled and unlabeled samples.  
Denoted as SSL-ART, the model leverages the family of ART neural networks that solve the stability-plasticity dilemma in online learning.  
Details of the proposed model is discussed in the following section.\par

\section{PRELIMINARIES}
\label{Sec:Pre}

\subsection{Fuzzy Subsethood Measure}
According to~\cite{kosko1986fuzzy}, the degree to which $\textbf{y}\in \mathbb{R}^D$ is a fuzzy subsethood of $\textbf{x}\in \mathbb{R}^D$ is given by:
\begin{align}
\label{eq:fsh}
    T=\frac{|\textbf{x}\wedge \textbf{y}|}{|\textbf{y}|},
\end{align}
where $\wedge$ represents the fuzzy $AND$ operator~\cite{Zadeh1965Fuzzy}, i.e.,
\begin{align} \label{eq:fuzzyy}
(\textbf{x}\wedge \textbf{y})\equiv min(x_i,y_{i}), i=1, ...,D,
\end{align}

and $|.|$ can be calculated as follows:
\begin{equation} \label{eq:norm}
|P|\equiv \sum_{i=1}^{r}|p_i|,
\end{equation}
where $r$ is the dimension of $P$.

In both fuzzy ART and fuzzy ARTMAP models, the choice function $T_j$~(\ref{eq:choic}) in the conservative limit (i.e., $\alpha=0$) reflects the degree to which the prototype node $W_j$ is a fuzzy subset of input $A$. 
As such, when
\begin{align}
\label{eq:fshh}
    \frac{|A\wedge W_j|}{|W_j|}=1,
\end{align}
then $W_j$ is said to be a fuzzy subset choice for input $A$.

Both fuzzy ART and fuzzy ARTMAP use the fuzzy subsethood measure~(\ref{eq:fsh}) to quantify the degree of similarity between their prototype nodes and input vectors.

\vspace*{-0.28cm}
\subsection{Fuzzy ART}
\label{Sec:sec:FART}

\begin{figure}[tb]
 \begin{center}
 \includegraphics[scale=0.6]{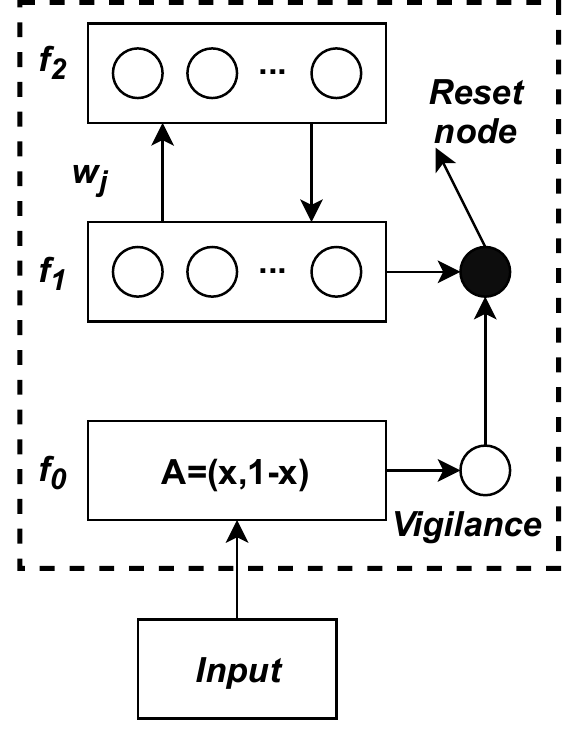}
  \end{center}
  \vspace*{-0.2cm}
  \caption{The structure of the fuzzy ART network. 
  The normalization layer ($f_0$) layer receives the input sample and performs complement-coding before propagation to the input layer ($f_1$).  
  The similarity level, i.e., Eq.~(\ref{eq:choic}), between the complement-coded input sample and each prototype node in the recognition layer ($f_2$) is computed, and the best-matched node is selected for the vigilance test using Eq.~(\ref{eq:vig_test}).  
  If none of the existing prototype nodes in $f_2$  can fulfil the condition in Eq.~(\ref{eq:vig_test}), a new prototype node is created in $f_2$ to encode the input sample.}
  \label{fig:FART}
\end{figure}

Fig.~\ref{fig:FART} shows the structure of fuzzy ART. 
It consists of three layers: normalization ($f_0$), input ($f_1$) and recognition ($f_2$) layers.
The complement-coding is used in the normalization layer to pre-process the input data.
This pre-processing step avoids the problem of category proliferation~\cite{Carpenter1991Fuzzy} by transforming a $D$-dimensional input sample, i.e., $\textbf{x}=(x_1,...,x_{D})$, into a $2D$-dimensional sample, as follows:
\begin{equation} \label{eq:comp}
A=(\textbf{x},1-\textbf{x})=(x_{1},...,x_{D},1-x_1,..., 1-x_{D}).
\end{equation}

The complement-coded sample received by the input layer is propagated to the recognition layer. The recognition layer is a dynamic layer which contains prototype nodes, and new prototype nodes can be added whenever necessary.

During learning, fuzzy ART computes the similarity level between the current input sample (\emph{A}) and the \emph{j}th prototype node stored in $f_2$ using choice function, as follows:
\begin{equation} \label{eq:choic}
T_{j}=\frac{|A\wedge W_{j}|}{\alpha+|W_{j}|},
\end{equation}
where $W_j= (w_{j,1},...,w_{j,2D})$ is the weight vector of the \emph{j}th node in $f_2$, which is initialized as $(1,1,...,1)$, $\alpha>0$ is the choice parameter, $\wedge$ represents the fuzzy $AND$ operator~\cite{Zadeh1965Fuzzy}.

Fuzzy ART selects the best-matched prototype node as the winner (indicated by \emph{J}) as follows:
\begin{equation} \label{eq:win}
 J=\underset{j=1,..., N}{\arg\max}~~T_{j},
\end{equation}
where \emph{N} is the number of prototype nodes in $f_2$. Resonance occurs if the winning node \emph{J} satisfies the vigilance criterion:
\begin{equation} \label{eq:vig_test}
\frac{|A\wedge W_{J}|}{|A|} > \rho,
\end{equation}
where $\rho$ is the user-defined vigilance parameter.
However, a mismatch occurs if the condition in~(\ref{eq:vig_test}) is not satisfied. Then, the current winning node $J$ is de-activated ($T_J=0$), and a new search cycle is triggered to select a new winning node.
This search cycle continues until one of the existing prototype nodes in $f_2$ is able to satisfy the criterion in~(\ref{eq:vig_test}). If no such node exists, a new prototype node is created in $f_2$ to encode the current input.

Finally, learning takes place, in which the weight vector of the winning node \emph{J} in $f_2$ is updated as follows:
\begin{equation} \label{eq:learn}
W_J^{(new)}=\beta(A\wedge W_J^{(old)})+(1-\beta)W_J^{(old)},
\end{equation}
where $\beta$ is the learning rate.

\subsection{Fuzzy ARTMAP}
\label{sec:sec:FAM}

\begin{figure}[tb]
 \begin{center}
 \includegraphics[scale=0.55]{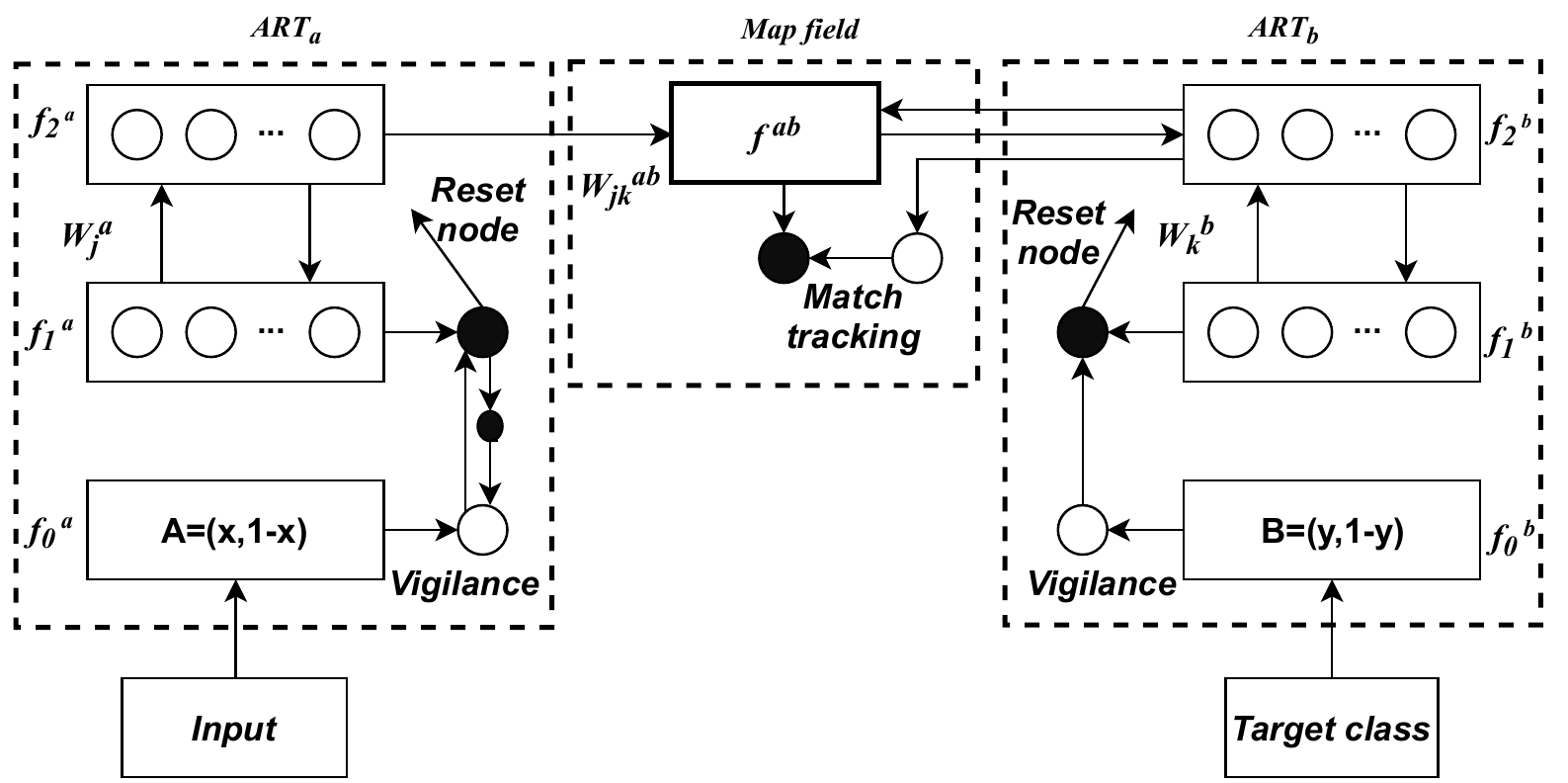}
  \end{center}
  \vspace*{-0.12cm}
  \caption{The structure of the fuzzy ARTMAP network. 
  It has two fuzzy ART models, i.e., $ART_a$ and $ART_b$, and a map field. 
  $ART_a$ and $ART_b$ receive the input sample and its corresponding target, respectively. 
  After identifying the winning prototype nodes in $ART_a$ (denoted as node $J$) and $ART_b$ (denoted as node $K$), the map field vigilance test (i.e., Eq.~(\ref{eq:map_field})) is used to map the $J$-th node in $f_a^2$ to the $K$-th node in $f_b^2$. If the condition in Eq.~(\ref{eq:map_field})) is not satisfied, a new search cycle is triggered for learning the current input sample. }
  \label{fig:FAM}
\end{figure}

As shown in Fig.~\ref{fig:FAM}, fuzzy ARTMAP consists of two fuzzy ART models, i.e., $ART_a$   and $ART_b$, and a map field ($f^{ab}$).
The complement-coded input sample (\emph{A}) and its corresponding target class (\emph{B}) are propagated to $ART_a$  and $ART_b$, respectively.
After selecting the winning nodes $J$ and $K$ in $f_2^a$ and $f_2^b$, respectively, the map-field vigilance criterion is applied:
\begin{equation} \label{eq:map_field}
\frac{|Y^b\wedge W_{j}^{ab}|}{|Y^{b}|}>\rho_{ab},
\end{equation}
where $\rho _{ab}$ is the map-field vigilance parameter, $W_j^{ab}$ is the weight vector from $f_2^a$ to $f^{ab}$, and $Y^b= (y_1^b,...,y_{N_b}^b)$ is the output vector of $f_2^b$:
\begin{equation}
y_l^{b}=
\begin{cases}
1, & if~~l=K\\
0, & otherwise
\end{cases}, l=1,...,N_b,
\end{equation}
where $K$ and $N_b$ indicate the winning node and number of nodes in $f_2^b$, respectively. If the criterion in~(\ref{eq:map_field}) is not satisfied, this means that the winning node \emph{J} in $f_2^a$ makes an incorrect prediction of the target class in $ART_b$.
In this case, a procedure known as match-tracking is triggered to correct the error.
During match-tracking, $\rho_a$ is updated to:
\begin{equation} \label{eq:match}
\rho_a=\frac{|A\wedge W_J^a|}{|A|}+\delta,
\end{equation}
where $\delta>0$.
As such, $ART_a$ ensues a new search cycle with the updated $\rho_a$ to find another winning prototype node (potentially a new node) that is able to satisfy~(\ref{eq:map_field}).
When this occurs, learning takes place to update  the weight vector of the winning node \emph{J} in $f_2^a$ using~(\ref{eq:learn}).

\vspace*{-0.3cm}
\vspace*{0.2cm}

\section{THE PROPOSED MODELS}
\label{Sec:Method}
This section, first, presents the proposed SSL-ART model with an OtM mapping scheme along with a numerical example and then discusses how to extract fuzzy \textit{If-Then} rules in order to explain the model's prediction. 
Finally, the weighted voting strategy along with the complexity analysis are presented.

\begin{algorithm}[tb!]
\caption{The learning phase of SSL-ART.}
\label{al:SSL-FART}
\KwIn{Parameters of SSL-ART ($\rho_a$, $\rho_b$, $\alpha_a$, $\alpha_b$, $\beta_a$, $\beta_b$), $D_U$ and $D_L$.\\
}
\KwOut{A trained SSL-ART model (prototype nodes of input features linked to the corresponding labels).\\
}
\For{each sample $\textbf{x}_u$ in $D_U$}{   
Perform complement-coding~(\ref{eq:comp});\\
Compute~(\ref{eq:choic}) for all nodes in $f_2^a$;\\
Select the prototype node $J$ with the highest choice value as winning node using~(\ref{eq:win});\\
Perform vigilance test~(\ref{eq:vig_test}) for node $J$;\\
\If {the vigilance test is not satisfied}{
Deactivate node $J$;\\
\If{all nodes in $f_2^a$ have been deactivated}{
Add a new node in $f_2^a$;\\
}\Else{
Go to step 2;
}
Update prototype node $J$ using~(\ref{eq:learn}) with respect to the current sample;\\
}
}

\For{each sample ($\textbf{x}_l,y_l$) in $D_L$ }{   
Find the winning prototype node $J$ in $f_2^a$ and update it with respect to $\textbf{x}_l$ using Steps 2-12; \\
Find the winning prototype node $K$ in $f_2^b$ and update it with respect to $y_l$ using Steps 2-12, by replacing superscript $a$ with $b$;\\
Update the OtM mapping of the winning node $J$ in $f_2^a$ to the winning node $K$ in $f_2^b$ using~(\ref{eq:OtM}). 
}
\end{algorithm}

\subsection{The Proposed SSL-ART Model}
\label{sec:sec:SSL-FART}
In this section, we propose a new two-stage SSL framework based on the ART models introduced in Section~\ref{Sec:Pre}.
\subsubsection{Learning} In the first stage, fuzzy ART is used to create a number of prototype nodes using unlabeled samples~$D_U= \{\textbf{x}_1,...,\textbf{x}_U\}$.
It starts with one new node (i.e. an uncommitted prototype) in $f_2^a$, $W_1=(w_{1,1},...,w_{1,2D})$, in which all the weight elements are initialized to 1. 
During learning, this new node is used to learn the patterns encoded in the training samples subject to meeting the vigilance criterion in~(\ref{eq:vig_test}). 
Once this prototype node is committed (i.e. used for weight update according to the learning rule), another new prototype node is dynamically created.\par

During the second stage, SSL-ART adopts the same network structure of fuzzy ARTMAP, but with a new OTM mapping strategy. 
Fuzzy ARTMAP comprises two fuzzy ART models, \textit{(i)} $ART_a$ for learning the input feature vectors, and \textit{(ii)} $ART_b$ for encoding the target classes. 
Given a small set of labeled samples~$D_L= \{(\textbf{x}_1,y_1 ),...,(\textbf{x}_L,y_L)\}$ in the second stage of SSL-ART, $ART_a$ is initialized with the prototype nodes created in the first stage using the unlabeled samples, and $ART_b$ starts with an uncommitted prototype node for learning the target classes. 
The fuzzy ARTMAP learning algorithm is used to map the input feature vectors in $ART_a$ to their respective target classes in $ART_b$.  
Specifically, given a pair of labeled data (input-target), $ART_a$ and $ART_b$, simultaneously receive the input feature vector and its target class, respectively.  
The best-matched node in $f_2^a$ (as well as in $f_2^b$) is selected as the winner using (\ref{eq:choic}) and (\ref{eq:win}). 
Next, the vigilance test (\ref{eq:vig_test}) is conducted to examine the similarity level.
In the event that all prototype nodes in $f_2^a$ fail to meet the vigilance criterion, a new node is dynamically created in $f_2^a$ to learn the association between the current pair of input feature vector and target class.\par

\subsubsection{The OtM mapping strategy}

Unlike fuzzy ARTMAP which forms a one-to-one (OtO) mapping in the map field, SSL-ART establishes a one-to-many (OtM) mapping strategy to associate each prototype node with one or more target classes.
To achieve this, SSL-ART updates the winning node \emph{J} using~(\ref{eq:learn}) without referring to the map-field vigilance criterion.
Suppose $F_j^{OtM}= (f_{j1}^{OtM},...,f_{jN_b}^{OtM})$, where $f_{jl}^{OtM}$ ($j=1,..., N_a$ and $l=1,...,N_b$) indicates the times that node $j$ in $f_2^a$ is associated with node $l$ in $f_2^b$, where $N_a$ and $N_b$ are the numbers of nodes in $f_2^a$ and $f_2^b$, respectively, and $F_j^{OtM}$ is initialized as $(0,...,0$). SSL-ART updates the OtM mapping of the winning node $J$ in $f_2^a$ to the winning node $K$ in $f_2^b$ as follows:
\begin{equation}\label{eq:OtM}
f_{Jl}^{OtM}=
\begin{cases}
f_{Jl}^{OtM}+1, & if~l=K\\
No~change, & otherwise
\end{cases}, l=1,...,N_b.
\end{equation}

This procedure continues for all the labeled samples.
{\color{black}Once all training samples are learned, SSL-ART maps each prototype node $j$ in $f_2^a$ to only one target class of node $k$ in $f_2^b$ that has accumulated the highest number of associations pertaining to all target classes, i.e.,:
\begin{align}\label{eq.14}
 k=\arg \max _{l=1,..,N_b}f_{jl}^{OtM}.   
\end{align} 

Eq.~(\ref{eq.14}) maps each prototype node in $f_2^a$ to only one target class in $f_2^b$ that contains the highest number of target samples, while other target classes are ignored, in order to tackle spurious and noisy training samples. 
}
The step-by-step learning phase of SSL-ART is summarized in Algorithm~\ref{al:SSL-FART}.\par

\subsubsection{Prediction}

To perform prediction, firstly, the similarity level~(\ref{eq:choic}) between the current test sample and all existing prototype nodes in $f_2^a$ is determined.
The winning node is selected using~(\ref{eq:win}).
Then, SSL-ART produces a prediction according to the winning node $J$ and~(\ref{eq:OtM}).
If the winning node \emph{J} is not linked to any target class, SSL-ART selects the next best node in $f_2^a$ to provide a prediction.
This cycle continues for \emph{T} best nodes in $f_2^a$.
If none of the \emph{T} nodes are labeled, SSL-ART is unable to yield prediction for the current test sample.\par

\begin{figure}
\centering
    \begin{subfigure}[b]{0.24\textwidth}
            \includegraphics[width=\textwidth]{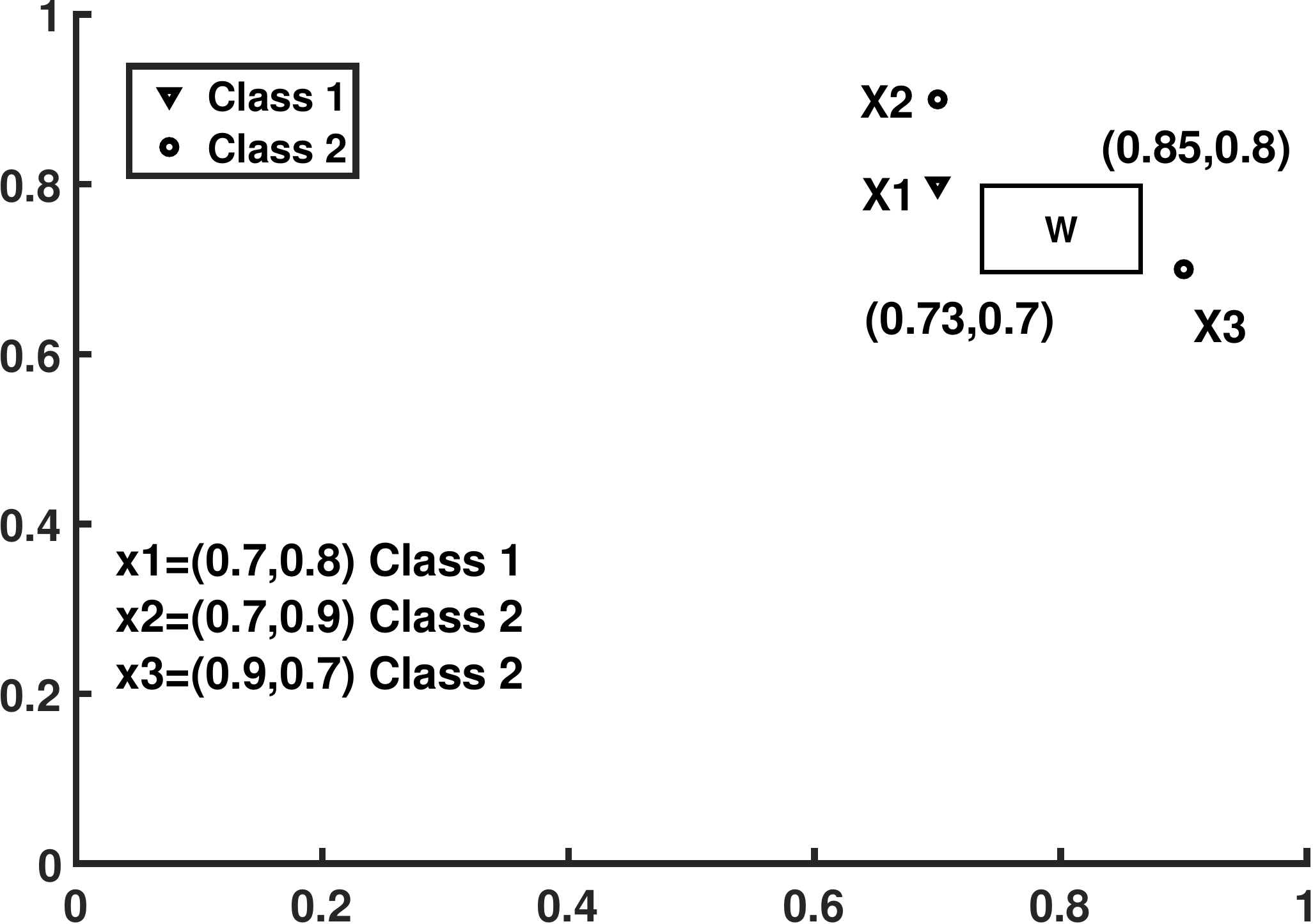}
            \caption{}
            \label{fig:SSL_F1}
    \end{subfigure}%
    \begin{subfigure}[b]{0.24\textwidth}
            \centering
            \includegraphics[width=\textwidth]{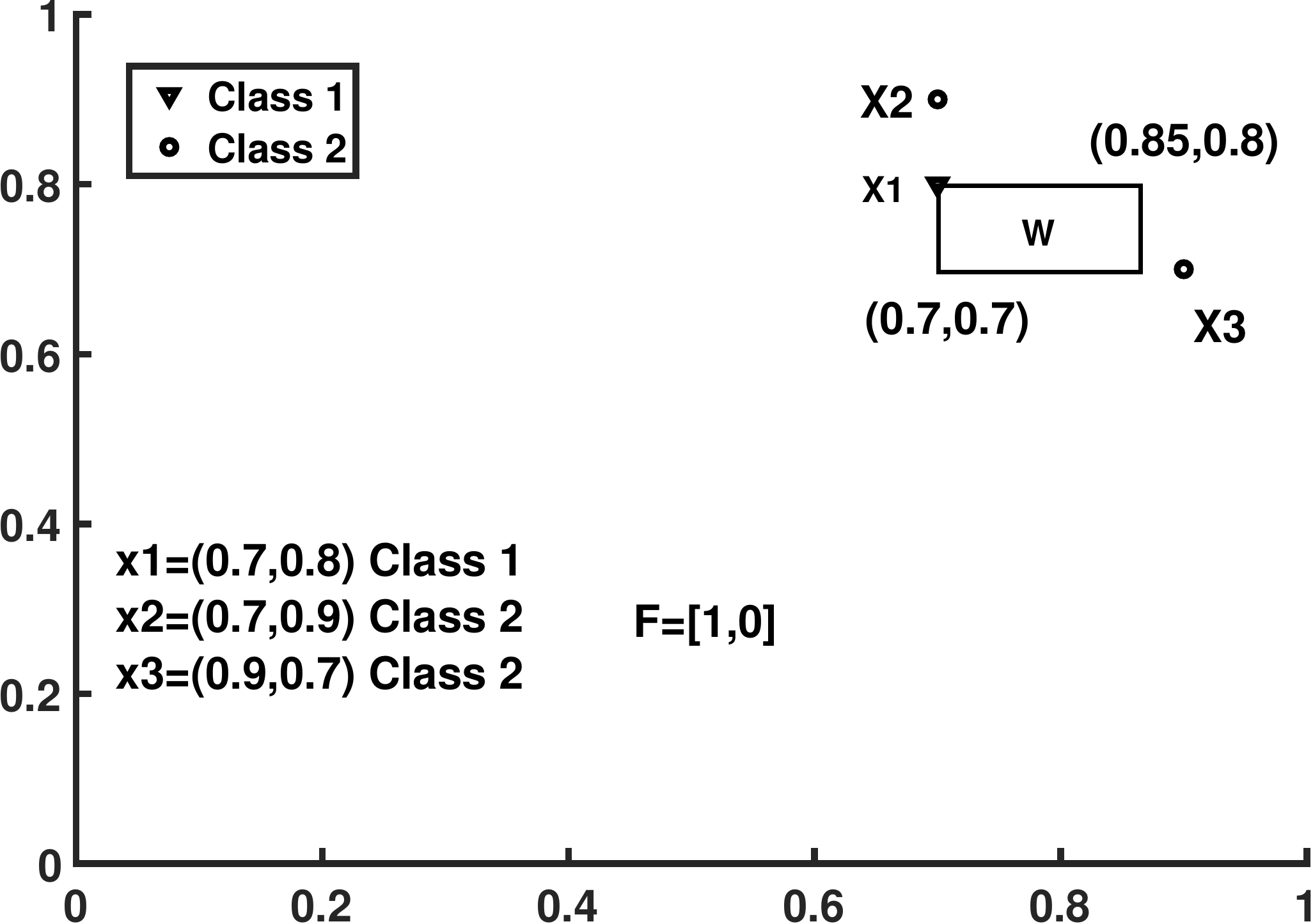}
            \caption{}
            \label{fig:SSL_F2}
    \end{subfigure}\\
    \vspace*{0.5cm}
    \begin{subfigure}[b]{0.24\textwidth}
            \centering
            \includegraphics[width=\textwidth]{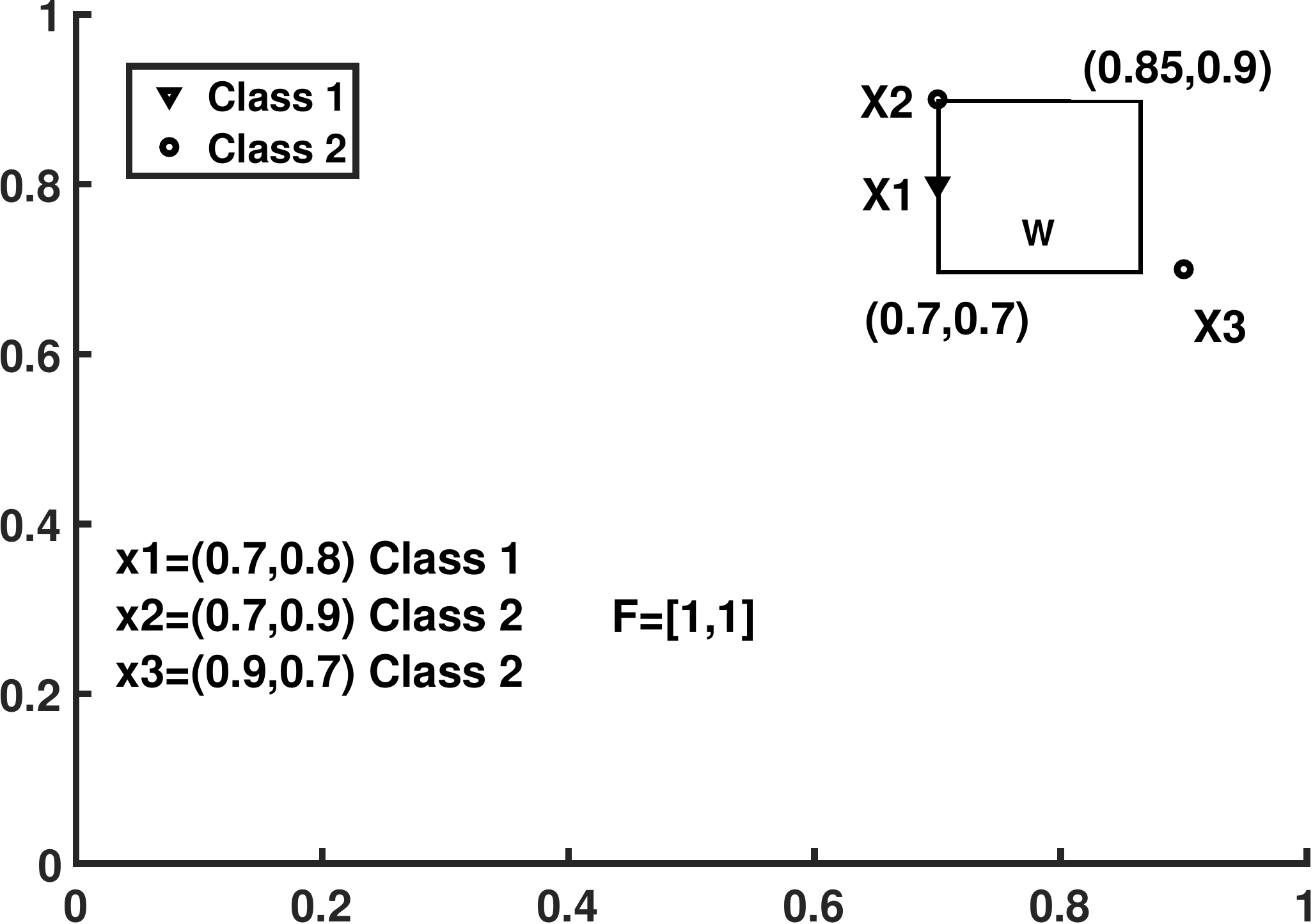}
            \caption{}
            \label{fig:SSL_F3}
    \end{subfigure}
    \begin{subfigure}[b]{0.24\textwidth}
            \centering
            \includegraphics[width=\textwidth]{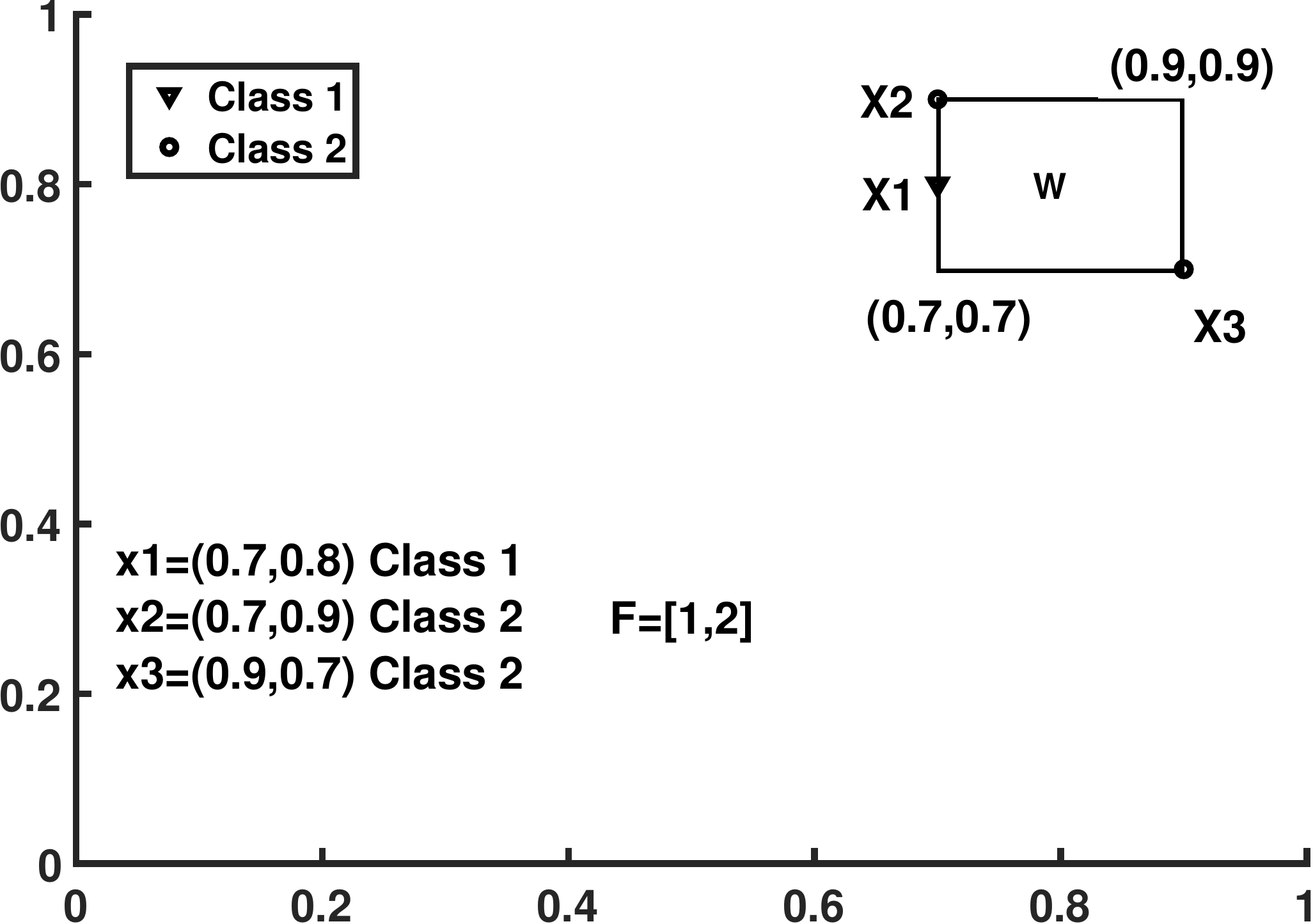}
            \caption{}
            \label{fig:SSL_F4}
    \end{subfigure}
    \vspace*{0.5cm}
    \caption{The learning phase of SSL-ART.}\label{fig:ComProb}
\end{figure}

\subsection{Discussion and Numerical Example }
\label{sec:sec:Dis}
The OtM mapping in SSL-ART has several advantages: it \emph{(i)} reduces the dependency of SSL-ART to the sequence of learning samples, \emph{(ii)} reduces the effects of noisy samples during the learning process; \emph{(iii)} avoids the overfitting problem by creating fewer prototype nodes. These can be seen in the following example.

A diagram indicating the learning phase of SSL-ART is shown in Fig.~\ref{fig:ComProb}.
Assume $W$ is an unlabeled node in $f_2^a$, and $\textbf{x}_1$, $\textbf{x}_2$ and $\textbf{x}_3$ are three labeled samples that belong to classes 1, 2 and 2, respectively, and $\textbf{x}_1$ is a noisy sample.
Firstly, SSL-ART receives $\textbf{x}_1$, and selects $W$ as the winning node (using~(\ref{eq:choic}), (\ref{eq:win}) and (\ref{eq:vig_test})).
Then, as shown in Fig.~\ref{fig:ComProb}~(b), $W$ and $F^{OtM}$ are updated based on~(\ref{eq:learn}) and~(\ref{eq:OtM}), respectively. Note that $F$ in Fig.~\ref{fig:ComProb} indicates $F^{OtM}$.
This procedure is repeated for $\textbf{x}_2$  (Fig.~\ref{fig:ComProb}~(c))  and $\textbf{x}_3$  (Fig.~\ref{fig:ComProb}~(d)).
Once all samples are learned, SSL-ART labels $W$ as class 2.
However, if we use the standard fuzzy ARTMAP model to learn these samples, according to the sequence of the learning samples, $W$ could be labeled either as class 1 or class 2 when the first sample comes, and it needs to create one or two more nodes to learn all samples.
While the OtM mapping in SSL-ART is able to reduce the effects of noisy samples, it is subject to the lose of information in decision boundary pertaining to different classes.\par

\vspace{-0.35cm}

\subsection{Extracting Fuzzy if-then Rules}
\label{Sec:sec:FAMART}
To extract fuzzy rules, each generated prototype node is considered as a rule describing the properties of a cluster of data samples.  
For rule interpretation, the input features are quantized into $Q$ levels according to the mechanism proposed in~\cite{carpenter1995rule}.  
As an example, let $Q$ be the number of quantized partitions for each fuzzy membership function. 
If $Q=5$, each feature can be interpreted in five quantized levels, e.g. ``Very small”, ``small”, ``Medium”, ``Large”, and ``Very large”.  
The round-off method is used for quantization by dividing the interval of [0,1] into $Q$ partitions as follows~\cite{carpenter1995rule}:
\begin{equation} \label{eq:quan}
V_q=\frac{q-1}{Q-1}
\end{equation}
where $q=1,...,Q$.

The extracted fuzzy rule $R_j$ can be written in the following form:\\
$R_j:~\textbf{if}~x_{p1}~is~V_q~...~\textit{and}~x_{pD}~is~V_q,\\ ~~~~~~~~~~~~\textbf{then}~\textbf{x}_p~\textit{belongs}~\textit{to}~C_i~\textit{with}~\textit{confidence}~p_i.$\\
where $\textbf{x}_p=(x_{p1},...,x_{pD})$ is a $D$-dimensional test sample and $V_q$ is the antecedent value.\par

\begin{table}[!t]
\centering
\caption{Details of the UCI data sets.}
\label{Table:uci}
    \begin{tabular}{l c c c}
    \toprule
    Data set & No.         & No.          & No.  \\
             & samples     & features     & classes\\
      \midrule
      Australian (AUS)              & 690 & 14 & 2\\
      Bupa                          & 345 & 6  & 2\\
      German (GC)                   & 1000& 24 & 2\\
      HaberMan's Survival (HMS)   & 306 & 3 & 2\\
      Heart Disease Cleveland (HDC) & 303 & 13 & 2\\
      Ionosphere (ION)              & 351 & 34 & 2 \\
      Kr-vs-Kp (KVK)                & 3196 & 36 & 2\\
      Mammographic Mass (MM)        & 961 & 5 & 2\\
      Pima Indians Diabetes (PID)   & 768 & 8 & 2\\
      WDBC                          & 569 & 30 & 2\\
      Wine                          & 178 & 13 & 3\\
      Iris                          & 150 & 4  & 3\\
      Seeds                         & 210 & 7  & 3\\
      Zoo                           & 101 & 16 & 7\\
      Optdigits      & { 5620}& { 64} & { 10}\\
      USPS      & { 9298}& { 256} & { 10}\\
      NORB       & { 48600}& { 2048} & { 5}\\
      MNIST       & { 70000} & { 784} & { 10}\\
      \bottomrule
     \end{tabular}
\end{table}

{  Note that SSL-ART is not a Mamdani fuzzy system. 
In a Mamdani fuzzy system, the inference procedure is performed using a set of $If-Then$ rules extracted from human experts.  
Comparatively, the $If-Then$ rules in SSL-ART are extracted based on the hyperbox structures learned from data samples.
\subsection{Weighted Ensemble SSL-ART Model}
\label{sec:sec:ESSL-FART}

Since each SSL-ART model trained with a different sequence of training samples can produce a different prediction, a pool of SSL-ART models can be created to form an ensemble model. 
The aim is to mitigate the effects of the sequence of training samples during online learning of each SSL-ART.  
As such, a weighted ensemble SSL-ART model (WESSL-ART) is proposed. Each ensemble member (i.e., SSL-ART) assigns a weight to each class based on its performance metric pertaining to a validation data set, as follows:
\begin{align}\label{eq:wei}
    We_{c}^m = \frac{NCS_{c}^m}{TNS_{c}},
\end{align}
where $We_{c}^m$ indicates the weight that ensemble member $m$ ($m=1, ..., M$) assigns to class $c$ ($c= 1,...,C$), $NCS_{c}^m$ is the number of samples that ensemble member $m$ has correctly classified as class $c$, $TNS_{c}$ is the total number of samples belonging to class $c$.
Using~\ref{eq:wei}, $C$ weights are computed by each ensemble member $m$.\par

During the test phase, each ensemble member yields a prediction for each test sample, $x$, along with the predicted class weight, $Weight(x)_{c}^m$, as follows:
\begin{equation}
Weight(x)_{c}^{m}=
\begin{cases}
We_{c}^m, & if~p^m(x)=c\\
0, & otherwise
\end{cases},
\end{equation}
where $p^m(x)=c$ indicates that class $c$ is the predicted outcome of test sample $x$ from ensemble member $m$.  
Then, the prediction score (PS) for each class pertaining to $x$ is computed, as follows:
\begin{align}
    PS_{c} = \sum_{i=1}^{M}Weight(x)_{c}^{m}.
\end{align}

The final outcome is the target class with the highest prediction score.

}

\subsection{Complexity Analysis}
\label{Sec:sec:com}
In this section, the Big-O notation~\cite{cormen2009introduction} is used to analyze the computational complexity of the SSL-ART model.
Let $D_a(D_b)$ and $N_a(N_b)$ be the numbers of features and prototype nodes in $ART_a (ART_b)$, respectively.
As reported in~\cite{burwick1998optimal}, the worst-case time complexity of fuzzy ART is $N_a^2+N_aD_a$, which is equivalent to $O(N_a^2)$ when $N_a\to \infty$ and $D_a\to \infty$.
Based on this reported outcome, the worst-case time complexity of SSL-ART consists of $(N_a^2+D_aN_a)+(N_a^2+D_aN_a)+(N_b^2+D_bN_b)+(N_b)$, which correspond to $ART_a$ in the first stage, $ART_a$ in the second stage, $ART_b$ in the second stage, and OtM mapping with~(\ref{eq:OtM}), respectively. This yields $O(N_a^2 +N_b^2)$ when $N_a\to \infty$, $D_a\to\infty$, $N_b\to \infty$ and $D_b\to \infty$.
In other words, the computational complexity of SSL-ART hinges on the number of prototype nodes in both $ART_a$ and $ART_b$.
Note that $N_b$ is the same as the number of target classes in classification problems, while $N_a$ (which is more influential) is controllable by the vigilance parameter, $\rho_a$.
We can set $\rho_a$ to a small threshold so that the time complexity of SSL-ART is within an acceptable duration in handling large, real-world data sets.

\begin{figure}[tb!]
\centering
    \begin{subfigure}[t]{0.24\textwidth}
            \includegraphics[width=\textwidth]{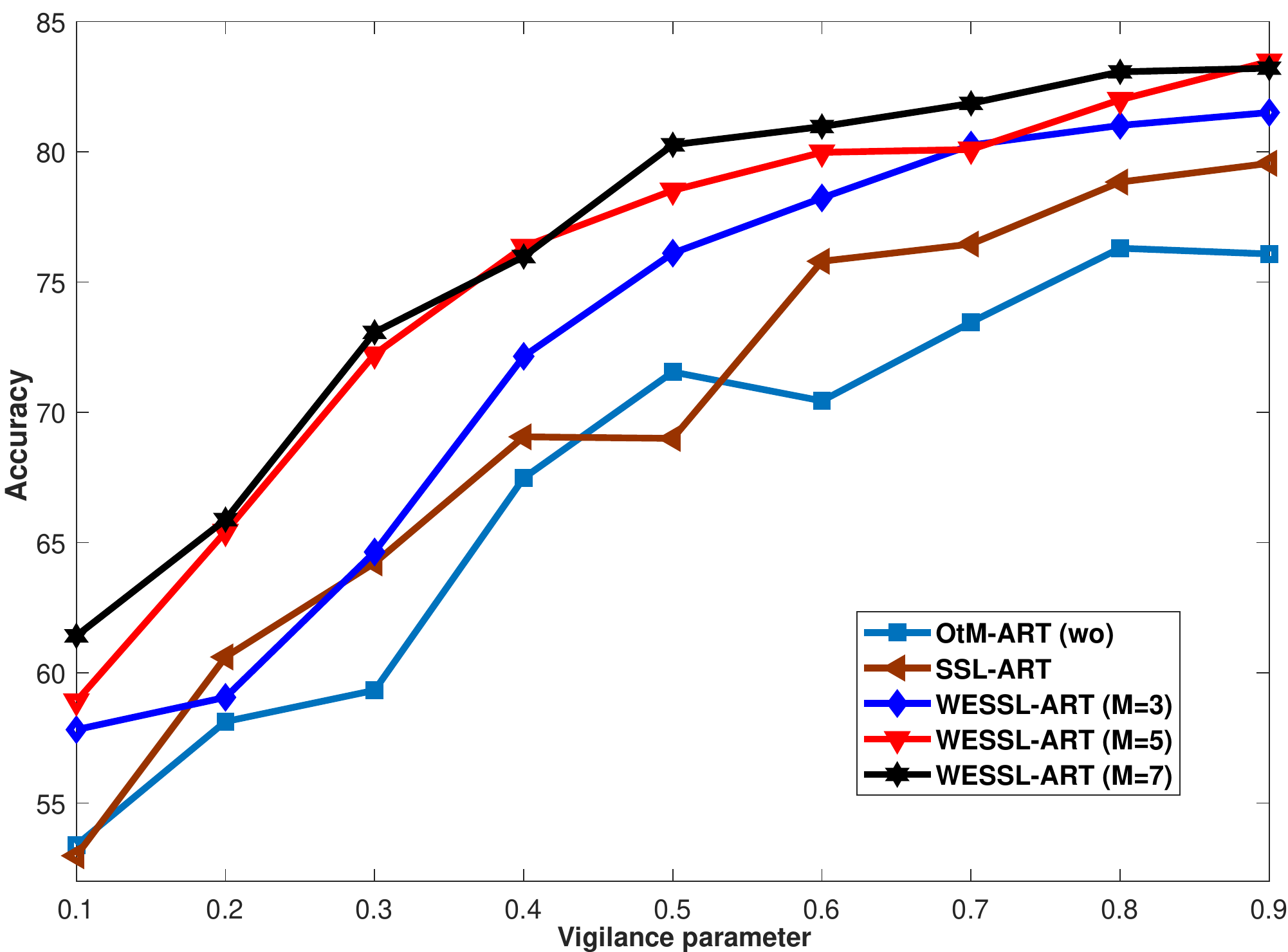}
            \caption{}
            \label{fig:HDC_Acc}
    \end{subfigure}
    \begin{subfigure}[b]{0.24\textwidth}
            \centering
            \includegraphics[width=\textwidth]{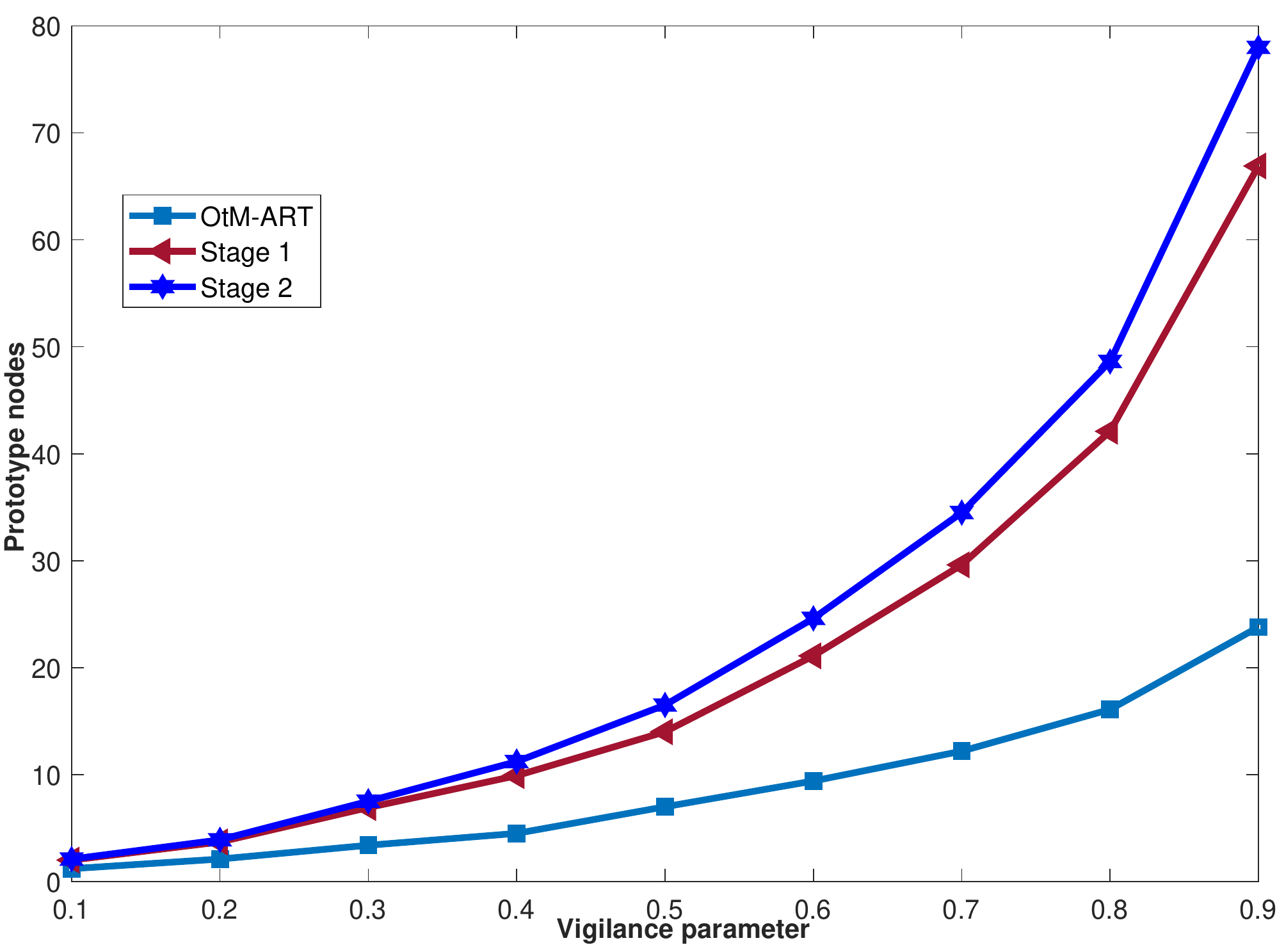}
            \caption{}
            \label{fig:HDC_Nodes}
    \end{subfigure}\\
    \vspace*{0.5cm}
    \begin{subfigure}[b]{0.24\textwidth}
            \centering
            \includegraphics[width=\textwidth]{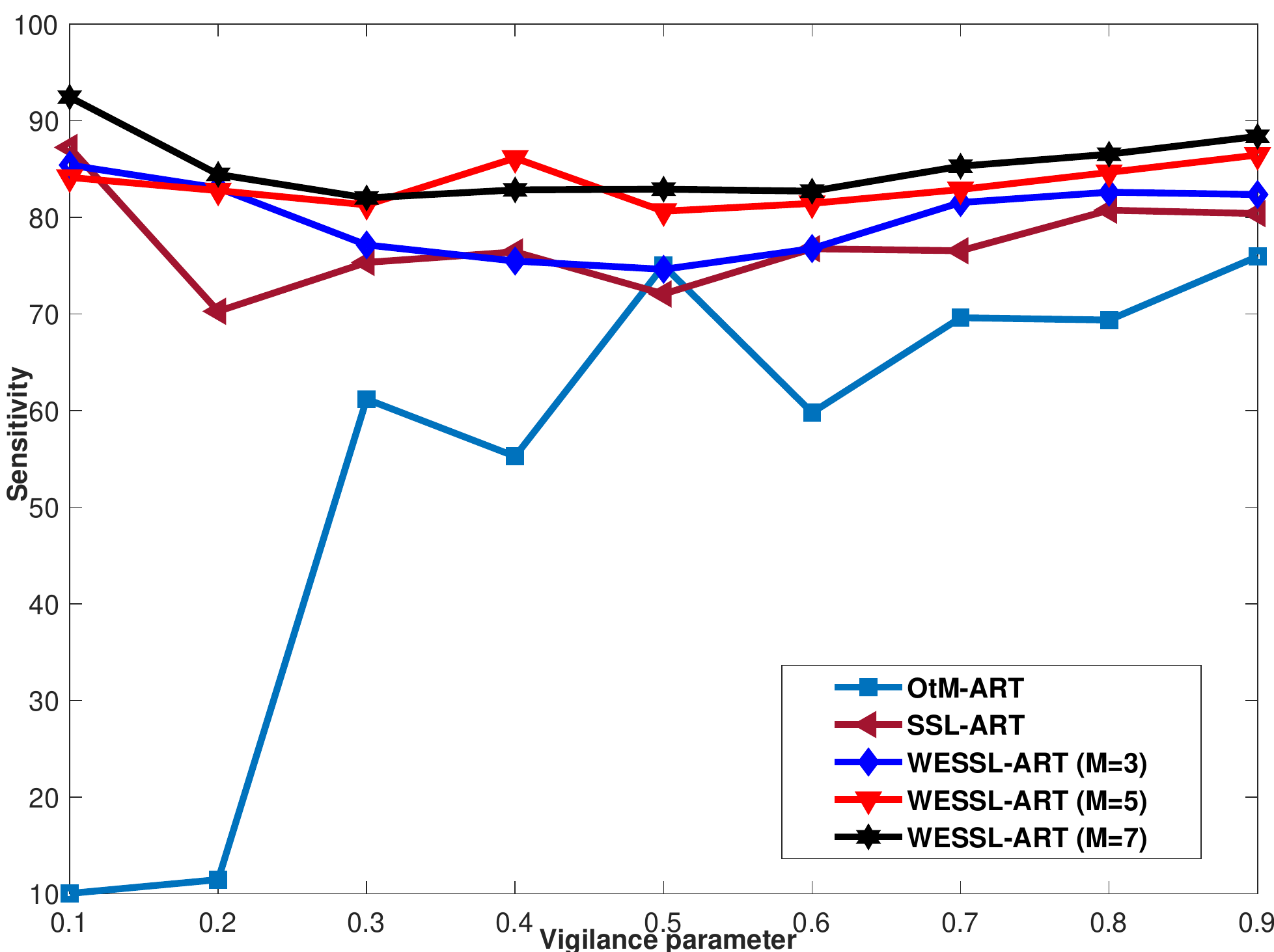}
            \caption{}
            \label{fig:HDC_Sens}
    \end{subfigure}
    \begin{subfigure}[b]{0.24\textwidth}
            \centering
            \includegraphics[width=\textwidth]{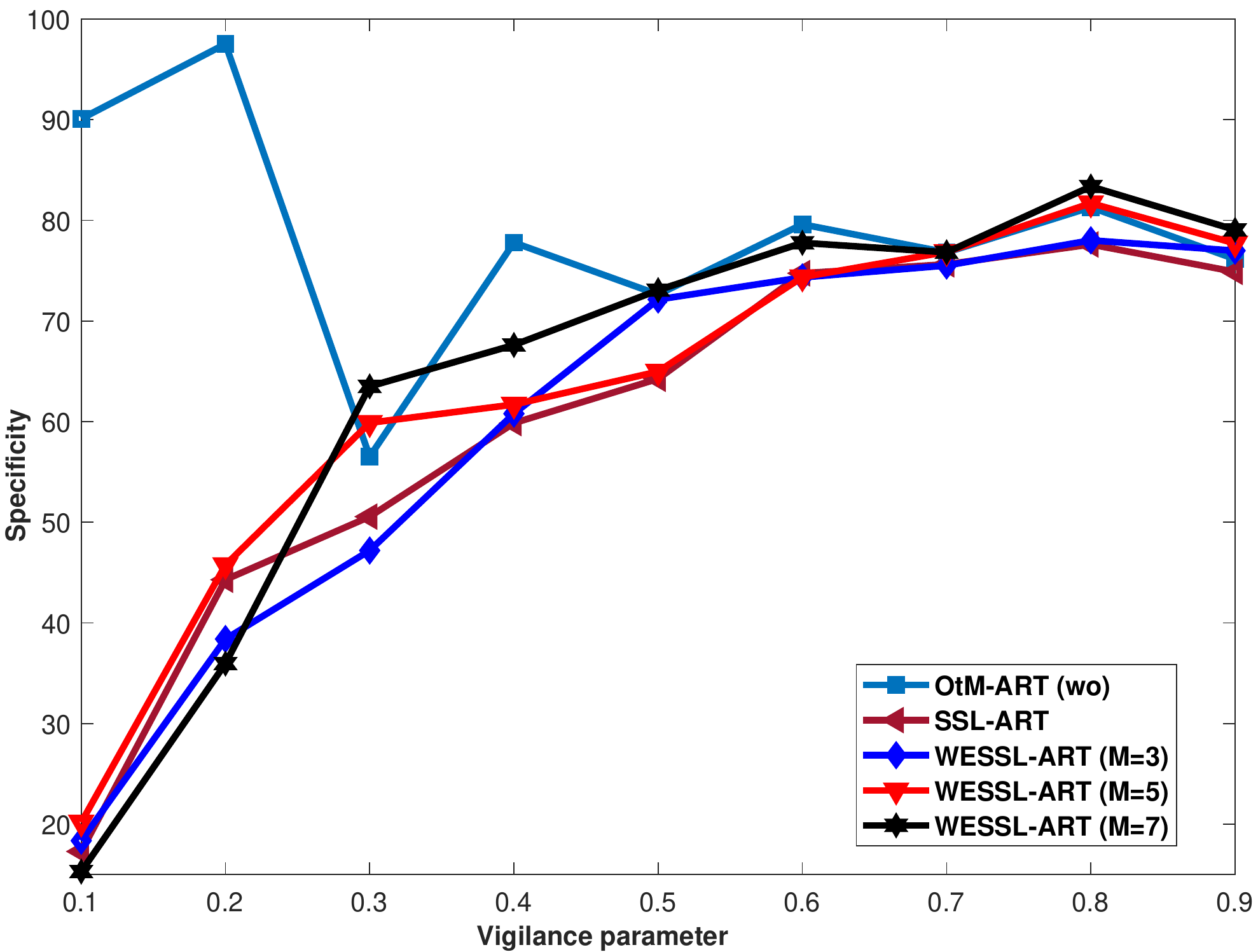}
            \caption{}
            \label{fig:HDC_Spec}
    \end{subfigure}
    \vspace*{0.2cm}
    \caption{  The effects of using unlabelled training samples in our proposed SSL-ART model  based on the HDC data set. (a) mean accuracy, and (b) the number of created prototype nodes. 
    Note that all models use OtM mapping and (wo) indicates that training is in a supervised manner without using unlabeled samples.
    }\label{fig:HDC}
\end{figure}

\section{EXPERIMENTAL STUDIES}
\label{Sec:ExP}
 \vspace*{-0.153cm}
Eighteen benchmark classification problems (Table~\ref{Table:uci}), three artificially generated data sets, and a real-world problem, i.e., human motion detection~\cite{pourpanah2019animproved}, have been used for performance evaluation.  Fourteen data sets are selected from the UCI machine learning repository~\cite{Dua2019UCI}, which contains small-scale data sets ranging from 101 to 3196 samples. Optdigits is medium-scale data set with 5500 samples that contain ten classes, and USPS, NORB, and MNIST are three large-scale image data sets with 10, 5, and 10 classes, respectively.
{\color{black}All features have been normalized between 0 and 1.}
For each data set, the experiment has been repeated 10 times.
In each repetition, all hyperparameters are kept constant, and only the training and test set samples are changed.
The bootstrap method has been applied to measure the mean accuracy scores along with their 95\% confidence intervals.
The parameters have been set as follows:~$\beta = \beta_a = \beta_b =1$, $\rho_b=1$, $\rho = \rho_a=[0.1,0.9]$, and the number of ensemble members ($M$) = 3, 5, and 7.\par

To have a fair performance comparison with other state-of-the-art models in Sections~\ref{sec:sec:UCI} and \ref{sec:sec:anomaly}, we have ensured that a prediction is made by SSL-ART for each test sample.
To achieve this, during the test phase, if the winning prototype node is unlabeled, the SSL-ART model continues searching for the next best prototype node (i.e., the highest $T_j$ measure) until a labeled prototype node is identified. 
While, in Section~\ref{sec:sec:Motion}, SSL-ART only conducts searching for the best (based on $T_j$) two ($T=2$) and three ($T=3$) prototype nodes in the neighborhood, to yield a prediction.
In the case none of them is labeled, SSL-ART is unable to yield prediction for the test sample.
The rational is to restrict the selection of the best-matched prototype nodes to two or three, such that a high-confident prediction is given for each test sample.
This leads to the tradeoff between correctness and coverage.
In this case, the coverage and correctness indicators of a classifier can be determined.
The coverage metric indicates the number test samples that have received a prediction over the total number of test samples, while the correctness metric is the number of correctly classified samples over the total number of samples that have received a prediction.
Finally, Section~\ref{sec:sec:rule} presents an analysis of the extracted rules.\par

\vspace{-0.359cm}
\subsection{Experiments with UCI Data Sets}
\label{sec:sec:UCI}
\subsubsection{Parameters setting}
In this experiment, the HDC data set has been used. 
The learning and test sets consist of 80\% and 20\% of the total data samples, respectively. 
The learning set has further been divided into 20\% labeled and 80\% unlabeled samples. 
To evaluate the effects of unlabeled samples on SSL-ART, we establish the {  OtM-ART (wo)} model, which only uses the labeled samples to build the network. 
In other words, the first stage of SSL-ART is omitted. 
Four performance indicators, i.e., accuracy, sensitivity, specificity, and number of prototype nodes have been computed.\par

\begin{figure}[tb!]
\centering
    \begin{subfigure}[b]{0.24\textwidth}
            \centering
            \includegraphics[width=\textwidth]{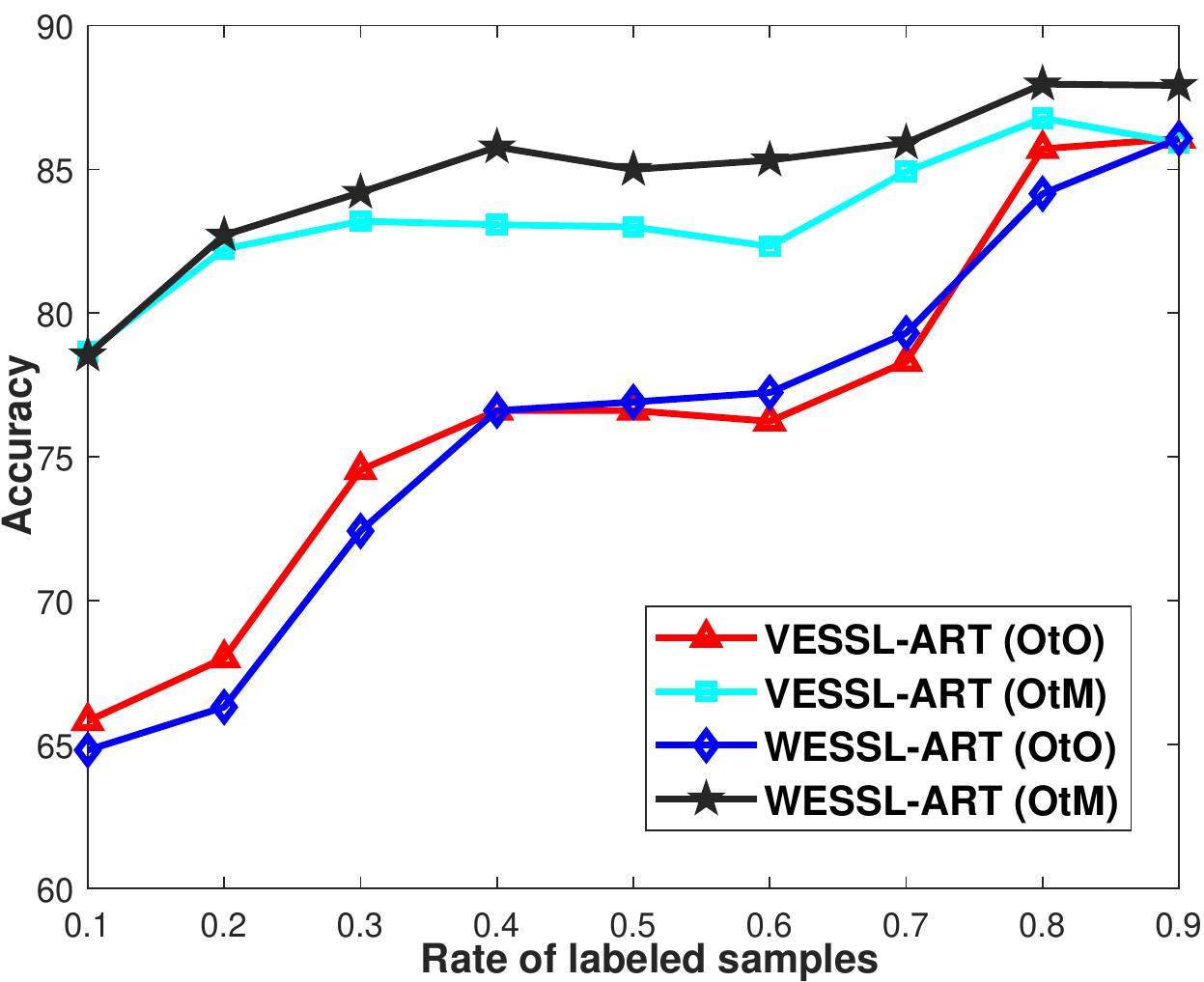}
            \caption{}
            \label{fig:HDC_acc1}
    \end{subfigure}
    \begin{subfigure}[b]{0.24\textwidth}
            \centering
            \includegraphics[width=\textwidth]{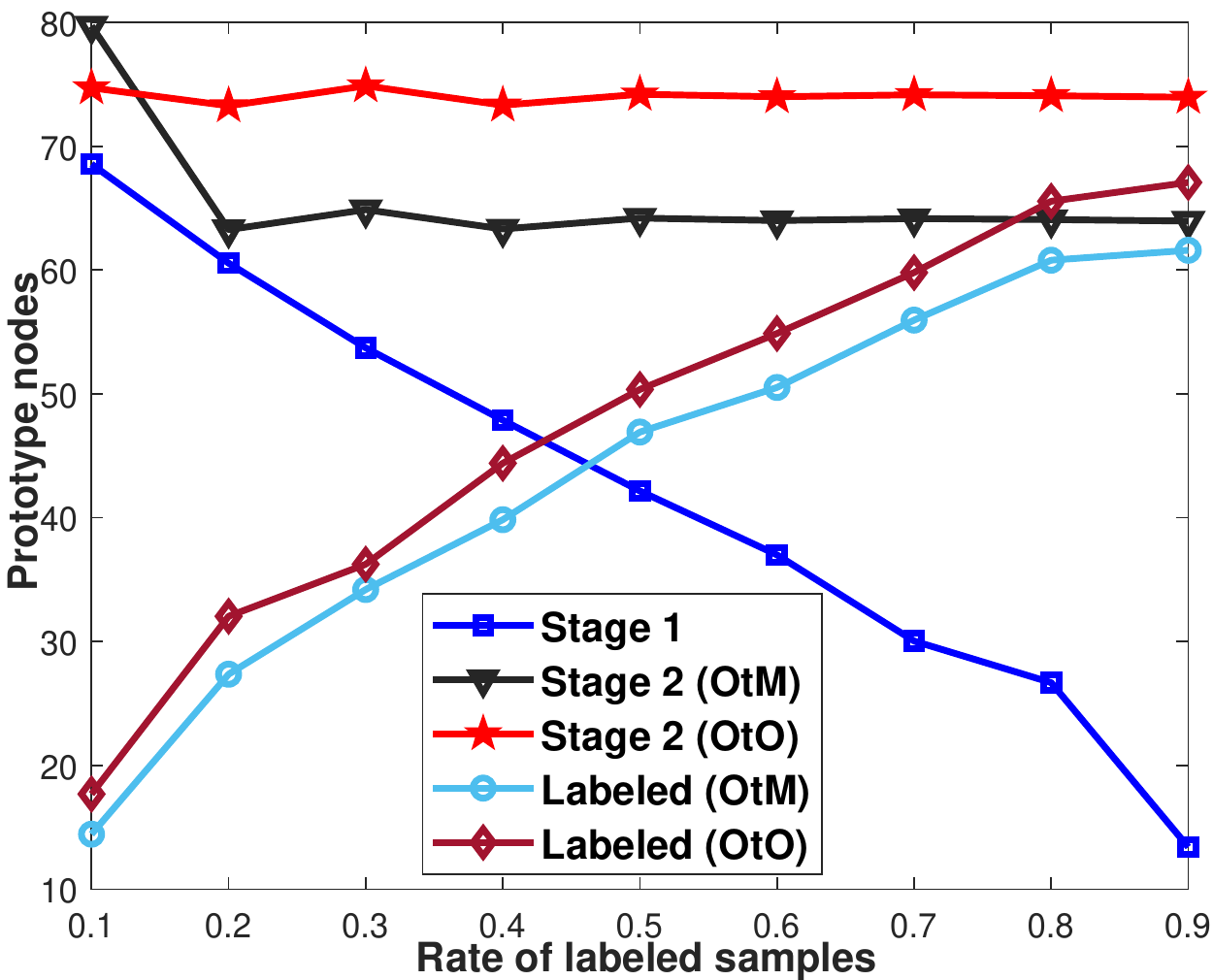}
            \caption{}
            \label{fig:HDC_node1}
    \end{subfigure}
    \vspace*{0.2cm}

    \caption{Performance comparison of OtM and OtO using different percentages of labeled samples based on the HDC data set. 
    (a) mean accuracy, and (b) number of created prototype nodes.}\label{fig:HDC_label}
\end{figure}

Fig.~\ref{fig:HDC}~(a) shows the accuracy rates of OtM-ART, SSL-ART, and { WESSL-ART} ($M$ = $3$, $5$ and $7$). The vigilance parameter has been varied from 0.1 to 0.9. {  It is noteworthy to mention that all models in Fig.~\ref{fig:HDC} are equipped with OtM mapping strategy. }
Overall, the accuracy rates of all models increase when the vigilance parameter is increased. { WESSL-ART} with $M=5$ and $7$ outperforms other models. 
{  Except for $\rho_a=0.5$, OtM-ART (wo) produces inferior results as compared with those of SSL-ART and WESSL-ART ($M=3$, $5$ and $7$), which shows the effects of the unlabeled samples in performance of the model.} Fig.~\ref{fig:HDC}~(b) shows the number of created nodes by {  OtM-ART (wo}), and SSL-ART (stage 1 and stage 2).
The complexity of both models increases when the vigilance parameter increases. In addition, SSL-ART adds more nodes in Stage 2 when the vigilance parameter increases from 0.3 to 0.9.
Unlike SSL-ART and { WESSL-ART} which exploit both unlabeled and labeled samples during the learning process, { OtM-ART (wo)} is not able to achieve a balanced sensitivity and specificity performance, as shown in Fig.~\ref{fig:HDC}~(c) and Fig.~\ref{fig:HDC}~(d), respectively.
This indicates the capability of the proposed model in extracting useful information from unlabeled samples.
To sum up, { WESSL-ART} is able to produce better results when $M=7$, and $\rho=0.9$. For the rest of the experiments, these values of $M$ and $\rho_a$ have been used for evaluation.\par

\begin{figure}[tb!]
\centering
    \begin{subfigure}[b]{0.24\textwidth}
            \centering
            \includegraphics[width=\textwidth]{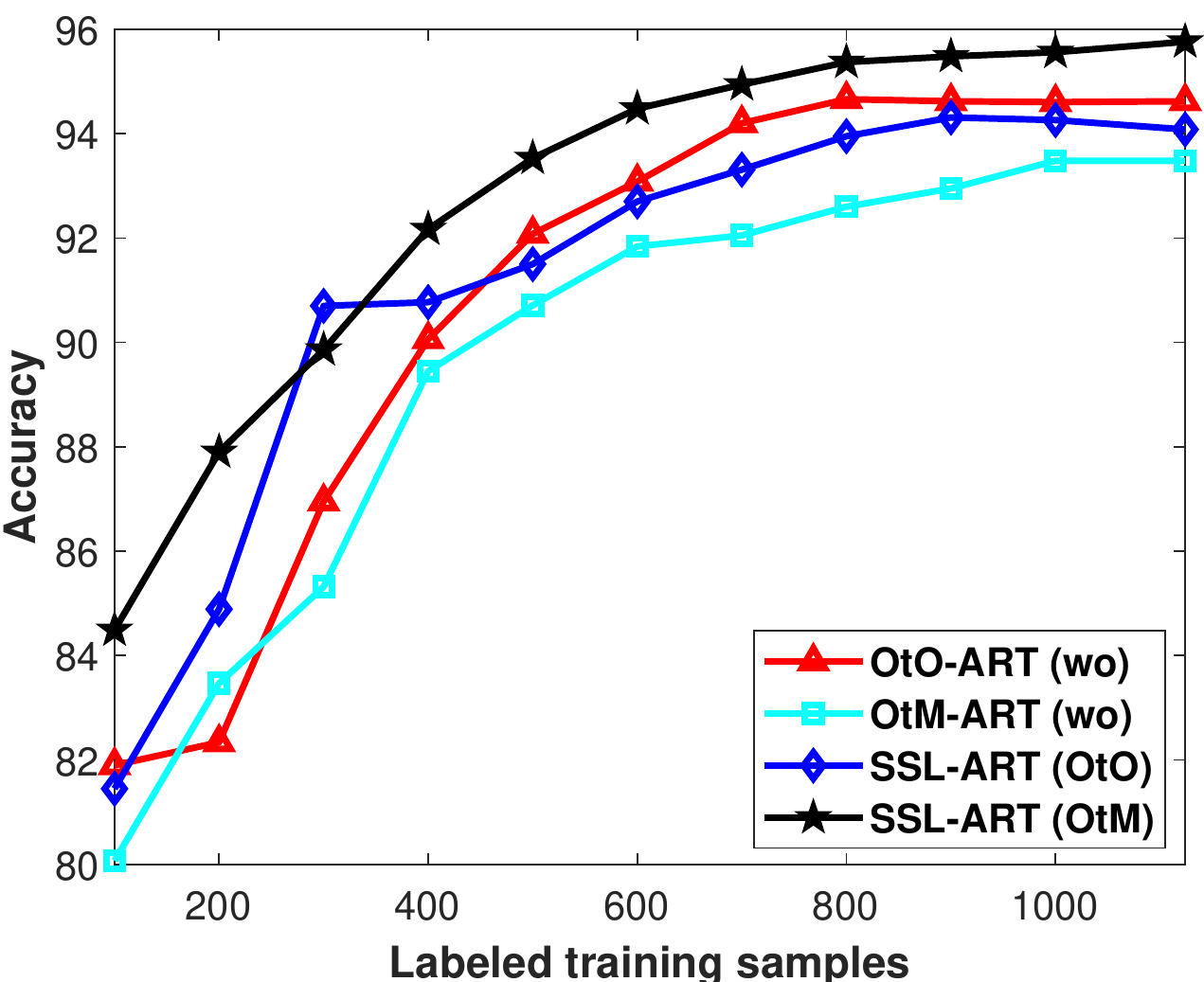}
            \caption{}
            \label{fig:acc1}
    \end{subfigure}
    \begin{subfigure}[b]{0.24\textwidth}
            \centering
            \includegraphics[width=\textwidth]{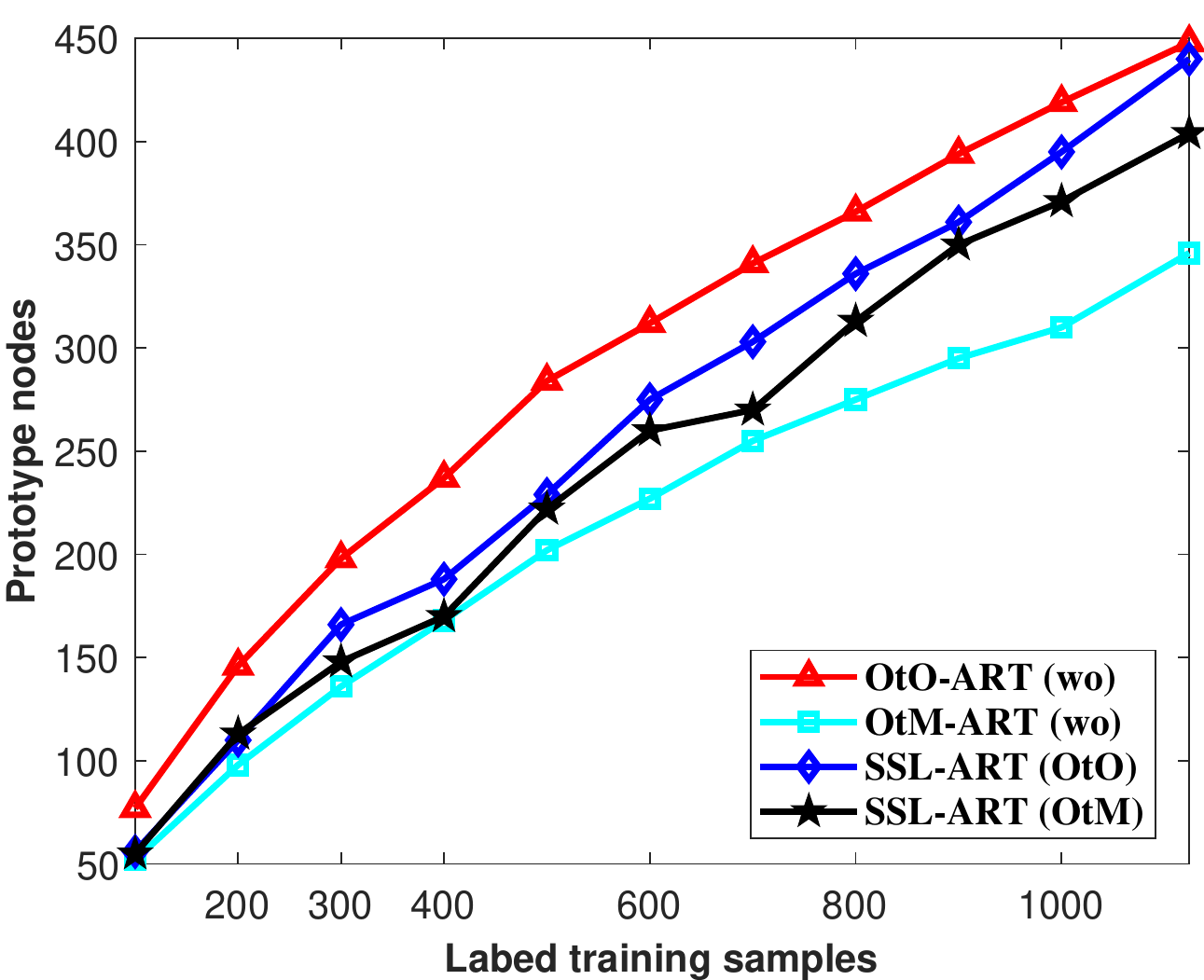}
            \caption{}
            \label{fig:node1}
    \end{subfigure}
    \vspace*{0.2cm}

    \caption{{ The (a) accuracy rates, and (b) a number of labeled prototype nodes of individually supervised and SSL models performing incremental learning using OtO and OtM mapping strategies using the Optdigits data set.  
    All models start learning with 100 labeled samples, and a cohort of 100 new labeled samples is added for incremental learning at each interval. 
    Note that (wo) indicates that training in a supervised manner without using unlabeled samples. } }\label{fig:inc}
\end{figure}

\subsubsection{OtM vs. OtO}
The aim of this experiment is two-fold: \textit{(i)} comparing the performance of one-to-many (OtM) mapping and one-to-one (OtO) mapping; \textit{(ii)} studying the effect of different percentages of labeled-unlabeled samples during the training process. 
To implement OtO mapping, the match-tracking function (\ref{eq:match}) is used to associate the generated prototype nodes with their corresponding target classes. 
Fig.~\ref{fig:HDC_label} (a) shows the accuracy rates of { WESSL-ART and VESSL-ART} ($M=7$) with both OtM and OtO mapping schemes based on the HDC data set. 
The accuracy rates of both mapping schemes increase when the percentage of labeled samples is increased from 10\% to 90\%. 
However, OtM mapping with both voting strategies produces stable results as compared with those of OtO mapping.\par

In term of the number of prototype nodes (Fig.~\ref{fig:HDC_label} (b)), both mapping schemes create the same number of unlabeled prototype nodes, as both use the same learning model (fuzzy ART) in their first stage. 
The numbers of prototypes nodes reduce when the percentages of unlabeled samples are decreased. 
However, in the second stage, the numbers of prototype nodes increase when the percentages of labeled samples increase. 
Overall, OtM mapping manages to produce fewer prototype nodes and tag them with the most likely labels (the highest number of cumulated target classes) as compared with those of OtO mapping. 
This is due to the match-tracking strategy (\ref{eq:match}) used by OtO mapping in associating each prototype node to only one target class, compromising the robustness in addressing noise in the training samples.\par

\begin{figure}[tb!]
 \begin{center}
 \includegraphics[scale=0.6]{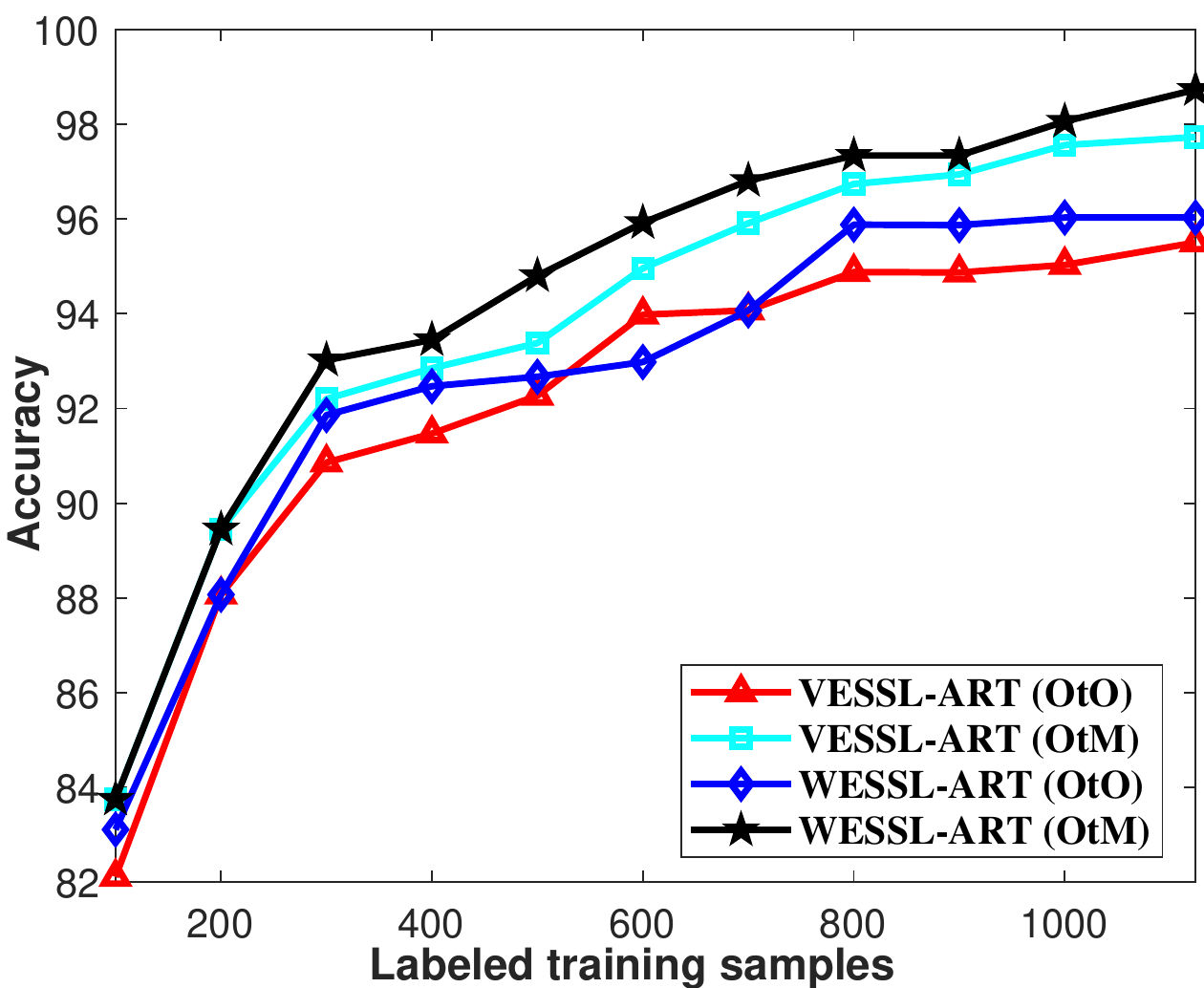}
  \end{center}
  \vspace*{-0.1cm}
  \caption{ The performance of WESSL-ART and VESSL-ART models with incremental learning using OtM and OtO mapping schemes for the Optdigits data set.}
  \label{fig:ens_inc}
\end{figure}

{ 
\subsubsection{Incremental learning}
This section studies the capability of incremental learning of the proposed SSL-ART model. 
Two experiments using the Optdigits data set are conducted.  
In both experiments, 90\% (60\% unlabeled and 20\% labeled) and 10\% of the data samples are used for training and test, respectively.  
All the evaluated models start with 100 labeled samples.  Then, a cohort of 100 new samples is added for incremental learning on an interval basis.\par

\begin{table}[tb] 
\vspace{0.9cm}
\centering \caption{\label{Table:sing}  Accuracy rates (Mean$\pm$SD) for USPS, NORB and MNIST data sets.}
    \begin{adjustbox} {width=\columnwidth}
    \begin{tabular}{l c c c c }
    \toprule
    Data set & SSL-ART (OtM) & FLAP-BLS        & TLLT-BLS        & LPDGL-BLS \\
    \midrule
    USPS &   87.10$\pm$1.91  & \hl{88.95$\pm$0.33} & 87.41$\pm$0.56 & 84.48$\pm$0.67 \\ 
    NORB &   \hl{$75.32\pm$2.21}  & 74.35$\pm$0.51 & 70.24$\pm$1.72 & 56.51$\pm$ 2.21\\ 
    MNIST &  \hl{87.76$\pm$1.46}  &85.84$\pm$0.33 & 80.27$\pm$0.46 & 83.94$\pm$0.51 \\ 
    \bottomrule
     \end{tabular}
    \end{adjustbox}
\end{table}

\begin{table*}[t] 
\centering \caption{\label{Table:Comp} { Accuracy rates $\pm$standard deviations of WESSL-ART, VESSL-ART, and SEMIB and REGB with three base classifiers for binary classification problems, along with numbers of win ($w$), tie ($t$) and loss ($l$).}}
    \begin{tabular}{l c c c c c c c c c}
    \toprule
      \multirow{2}{*}{Data set}& \multicolumn{2}{c}{Ensemble SSL-ART} & \multicolumn{2}{c}{Na\"{i}ve Bayes (NB)} & \multicolumn{2}{c}{C4.5 (DT)} & \multicolumn{2}{c}{3NN}\\
    \cmidrule{2-9}

                    & WESSL-ART & VESSL-ART & SEMIB~\cite{Mallapragada2008Semiboost} & REGB~\cite{Chen2011Semi} & SEMIB~\cite{Mallapragada2008Semiboost} & REGB~\cite{Chen2011Semi} & SEMIB~\cite{Mallapragada2008Semiboost} & REGB~\cite{Chen2011Semi}  \\
    \cmidrule{2-9}
     AUS            & \hl{87.18$\pm$1.7} & {86.34$\pm$1.4} & 78.70$\pm$2.6 & 81.70$\pm$3.0 & 82.20$\pm$5.1 & 84.50$\pm$3.4 & 64.90$\pm$3.8 & 62.00$\pm$4.1\\
    
     \emph{(w,t,l)} & (7,0,0)      &(6,0,1)&       (2,0,5)    & (3,0,4) & (4,0,3)   &  (5,0,2) & (1,0,6)   & (0,0,7) \\
      \cmidrule{2-9}
      Bupa      &\hl{67.04$\pm$3.8} & 65.89$\pm$2.0 & 56.50$\pm$7.7 & 61.60$\pm$6.5 & 55.70$\pm$4.5 & 61.70$\pm$5.4 & 58.30$\pm$4.5 & 59.10$\pm$5.9\\
    
     \emph{(w,t,l)} &(7,0,0)  & (6,0,1) & (1,0,6) & (4,0,3) & (0,0,7) & (5,0,2) & (2,0,5) & (3,0,4)\\
     \cmidrule{2-9}
     German    & 73.10$\pm$2.4 & 71.70$\pm$1.13  & 74.90$\pm$3.6 & \hl{75.60$\pm$3.3} & 72.00$\pm$2.7 & 73.20$\pm$3.4 & 70.20$\pm$3.1 & 67.60$\pm$3.6\\
    
     \emph{(w,t,l)} & (4,0,3)&(2,0,5)&(6,0,1) & (7,0,0) & (3,0,4) & (5,0,2) & (1,0,6) & (0,0,7)\\
     \cmidrule{2-9}
     HMS       & \hl{74.93$\pm$1.3} & {74.34$\pm$1.6} & 70.70$\pm$4.4 & 72.60$\pm$6.4 & 70.20$\pm$4.6 & 69.00$\pm$8.0 & 69.20$\pm$6.1 & 69.30$\pm$4.4\\
    
     \emph{(w,t,l)} &   (7,0,0)     & (6,0,1) &  (4,0,3) & (5,0,2) & (3,0,4) & (0,0,7) & (1,0,6) & (2,0,5)\\
     \cmidrule{2-9}
     HDC       & \hl{83.04$\pm$2.2} & {82.11$\pm$1.5} & 74.00$\pm$7.3 & 80.50$\pm$3.5 & 71.10$\pm$6.7 & 79.00$\pm$6.9 & 57.30$\pm$6.9 & 60.00$\pm$4.4\\
    
    \emph{(w,t,l)} &(7,0,0)  & (6,0,1) &   (3,0,4) & (5,0,2) & (2,0,5) & (4,0,3)  & (0,0,7) & (1,0,6)\\
    \cmidrule{2-9}
     ION       & \hl{93.02$\pm$2.8} & {91.20$\pm$2.6} & 82.30$\pm$3.8 & 90.10$\pm$4.7 & 82.30$\pm$6.9 & 85.70$\pm$4.8 & 74.60$\pm$7.4 & 80.10$\pm$6.8\\
    
     \emph{(w,t,l)} &   (7,0,0)     & (6,0,1)&      (2,1,4) & (5,0,2) & (2,1,4) & (4,0,3) & (0,0,7) & (1,0,6)\\
     \cmidrule{2-9}
     KVK       & 84.43$\pm$1.3 & 83.72$\pm$1.6 & 84.00$\pm$1.2 & 84.20$\pm$1.4 & 85.20$\pm$1.2 & \hl{87.80$\pm$0.8} & 73.80$\pm$1.6 & 78.10$\pm$1.4\\
    
     \emph{(w,t,l)} &   (5,0,2)    & (2,0,5) &      (3,0,4)   & (4,0,3)    & (5,0,2)   & (7,0,0)  & (0,0,7)   & (1,0,6)\\
     \cmidrule{2-9}
     MM        & \hl{84.37$\pm$3.1} & {82.84$\pm$1.7} & 80.30$\pm$3.6 & 80.60$\pm$3.5 & 80.70$\pm$1.3 & 80.50$\pm$3.2 & 77.60$\pm$3.3 & 74.10$\pm$3.2\\
    
     \emph{(w,t,l)} &   (7,0,0)     & (6,0,1) &      (2,0,5)        & (4,0,3)      & (5,0,2)       & (3,0,4)       & (1,0,6)       & (0,0,7)\\
     \cmidrule{2-9}
     PID       & \hl{77.45$\pm$21} & {76.15$\pm$2.0} & 72.20$\pm$3 & 74.20$\pm$4.4 & 72.50$\pm$3.9 & 72.90$\pm$4.1 & 70.30$\pm$4.1 & 69.60$\pm$4.1\\
    
     \emph{(w,t,l)}&    (7,0,0)    & (6,0,1) &       (2,0,5)   & (5,0,2)       & (3,0,4)       & (4,0,3)         & (1,0,6)     & (0,0,7)\\
     \cmidrule{2-9}
     WDBC      & 95.94$\pm$1.6 & \hl{96.18$\pm$1.6} & 94.30$\pm$2.1 & 94.40$\pm$2.9 & 92.10$\pm$2.9 & 95.30$\pm$1.8 & 89.60$\pm$3.6 & 80.10$\pm$2.2\\
    
     \emph{(w,t,l)}& (6,0,1)  & (7,0,0) &  (3,0,4) & (4,0,3) & (2,0,5) & (5,0,2) & (1,0,6) & (0,0,7)\\
     \cmidrule{2-9}
     \emph{(w,t,l)}&(64,0,6) &  (53,0,17) & (28,1,41) & (46,0,24) & (29,1,40) & (39,1,20) & (8,0,62) & (8,0,62)\\
    \bottomrule
     \end{tabular}
\end{table*}

The first experiment compares single SSL models in a supervised setting using both OtM and OtO mapping schemes. The experimental results are shown in Fig.~\ref{fig:inc}. Without utilizing unlabeled samples for training, OtM mapping, i.e., OtM-ART (wo), produces inferior results as compared with those from OtO mapping, i.e., OtO-ART (wo).  However, OtM mapping produces the best results when unlabeled samples are adopted for training.  In general, OtO mapping is not effective for SSL models, as shown in Fig.~\ref{fig:inc} (a).  
Fig.~\ref{fig:inc} (b) depicts the number of labeled prototype nodes in each setting. All models manage to add new prototype nodes whenever necessary, which is a key property of ART-based networks in incremental learning.\par

The second experiment compares the performance of weighted voting (WESSL-ART) with majority voting (VESSL-ART) using both OtM and OtO mapping strategies.  
As shown in Fig.~\ref{fig:ens_inc}, all ensemble models produce similar results when the number of labeled samples for training are fewer than 200. 
Then, weighted voting with OtM mapping, i.e., WESSL-ART (OtM), outperforms other models.  
In general, both weighted and majority voting strategies are able to yield better results when using SSL-ART with OtM mapping as their ensemble members.\par

\subsubsection{SSL-ART vs. other single SSL models} 

This experiment compares the performance between SSL-ART with OtM mapping and those of broad learning system (BLS)-based SSL models reported in~\cite{zheng2020novel} using three large-scale data sets, i.e., USPS, NORB, and MNIST data sets. Following the same experimental procedure in~\cite{zheng2020novel}, a total of 9298 samples are randomly selected from each data set, in which 9000 and 298 are training (8500 unlabeled and 500 labeled) and test samples, respectively. 
SSL-ART produces better results for the NORB and MNIST data sets, while the FLAP-BLS model has the best accuracy for the USPS data set, as shown in Table~\ref{Table:sing}.  
Again, SSL-ART requires only the setting of one (i.e. vigilance) parameter with one-pass learning through all the data samples.\par

}

\subsubsection{WESSL-ART vs. VESSL-ART and other ensemble models}

Two experiments have been conducted. 
In the first experiment, ten binary classification problems from the UCI machine learning repository have been used to compare { WESSL-ART and VESSL-ART} ($M=7$) with similar models reported in~\cite{Chen2011Semi}, i.e., SemiBoost (SEMIB)~\cite{Mallapragada2008Semiboost} and RegBoost (REGB)~\cite{Chen2011Semi}, with three base learners including $3$-nearest neighbors (3NN), Na\"{i}ve Bayes, and C4.5 Decision Tree.
Following the same procedure in~\cite{Chen2011Semi}, 80\% and 20\% of data samples have been used for learning and test, respectively.
For the learning data set, 20\% labeled and 80\% unlabeled samples have been used.\par

{ Table~\ref{Table:Comp} shows the mean accuracy rates with with standard deviations (SD) of WESSL-ART, VESSL-ART} and other methods, along with the numbers of win, tie and loss.
{ For all data sets except German, KVK and WDBC, WESSL-ART outperforms VESSL-ART and other SSL methods. For WDBC and KVK data sets, WESSL-ART is ranked as the second based method after VESSL-ART and REGB with C45 classifier, respectively. However, for the German data set, both WESSL-ART and VESSL-ART produce inferior results. In total, WESSL-ART and VESSL-ART, respectively, won 64 and 53 times, and lost only 4 and 17 times.}\par

The second experiment compares the performance of { WESSL-ART} and VESSL-ART with ensemble-based SSL methods, including MeanS3VM~\cite{Li2009Semi}, Laplacian SVM (Lap SVMp)~\cite{Belkin2006Manifold} and Multi-class Semi-Supervised Kernel Spectral Clustering (MSS-KSC)~\cite{Mehrkanoon2014Multiclass} for solving multi-class benchmark problems.
Four small data sets, i.e., Wine, Iris, Seeds, and Zoo, have been used. Following~\cite{Mehrkanoon2014Multiclass}, 80\% and 20\% of samples have been used for learning and test, respectively. Then, 25\% labeled and 75\% unlabeled learning samples have been formed.
The parameters of MeanS3VM and Laplacian SVM have been adopted from~\cite{Mehrkanoon2014Multiclass}. 
For MeanS3VM,  $C_1$ and $C_2$ have been set to 1 and 0.1, respectively. 
The accuracy rate of the validation set has been used to adjust the width parameter of the RBF kernel. 
For the Laplacian SVM model, $\gamma_I$ and NN have been set to 1 and 6, respectively, while the kernel parameter and $\gamma_A$ have been tuned according to the accuracy rate of the validation set.

Table~\ref{Table:EnsmCompare} shows the accuracy rates along with standard deviations.
{  WESSL-ART outperforms other methods for both the Iris and Seeds data sets, and produces similar results with that of VESSL-ART for the wine data set, and is ranked as the second best classifier for the Zoo data set.}\par

\begin{table}[tb]
\vspace{0.4cm}
\centering \caption{\label{Table:EnsmCompare} Accuracy rates (Mean$\pm$SD) for multi-class problems.}
    \begin{adjustbox} {width=\columnwidth}
    \begin{tabular}{l c c c c }
    \toprule
      Method & Wine &  Iris & Seeds & Zoo\\
    \midrule
    MSS-KSC~\cite{Mehrkanoon2014Multiclass}        & 0.96$\pm$0.02 & 0.89$\pm$0.08 & 0.90$\pm$0.04 & \hl{0.93$\pm$0.05}\\
    LAPSVMp~\cite{Belkin2006Manifold}        & 0.94$\pm$0.03 & 0.88$\pm$0.05 & 0.89$\pm$0.03 & 0.90$\pm$0.06\\
    Mean3vm-iter~\cite{Li2009Semi}   & 0.95$\pm$0.03 & 0.90$\pm$0.03 & 0.88$\pm$0.07 & 0.88$\pm$0.02\\
    Means3vm-mkl~\cite{Li2009Semi}   & 0.94$\pm$0.07 & 0.89$\pm$0.01 & 0.89$\pm$0.02 & 0.89$\pm$0.07\\
    VESSL-ART      & \hl{0.96$\pm$0.01} & 0.96$\pm$0.02 & 0.95$\pm$0.03 & 0.90$\pm$0.04\\
     WESSL-ART    &  { \hl{0.96$\pm$0.02}} &  { \hl{0.97$\pm$0.06}} &  { \hl{0.96$\pm$0.02}} &  { 0.92$\pm$0.02}\\
    \bottomrule
     \end{tabular}
    \end{adjustbox}
\end{table}

Moreover, Table~\ref{Table:node} summarizes the average numbers of prototype nodes created in the first and second stages, along with the numbers of prototype nodes that are labeled in the second stage by SSL-ART. 
Specifically, SSL-ART employs the (unlabeled) prototype nodes created at the first stage for further learning and labeling in the second stage with a small set of labeled training samples.  
New prototype nodes can be created when the existing ones are unable to learn the labeled input samples.\par

\begin{table}[t]
\vspace{0.4cm}
\centering \caption{\label{Table:node}The average numbers of prototype nodes created at each stage, along with the numbers of labeled prototype nodes by SSL-ART for the UCI data sets. }
    \begin{adjustbox} {width=\columnwidth}
    \begin{tabular}{l c c c |l c c c}
    \toprule
      Data set& Stage 1 & Stage 2 & Labeled & Data set& Stage 1 & Stage 2 & Labeled \\
    \midrule
     AUS            & 82.12 & 85.76& 51.23    &MM  & 74.07& 81.36 & 46.34\\
     Bupa      &47.50 & 55.50& 32.40          &PID       & 73.60 & 76.88& 56.34 \\
     German    &128.30&171.76 & 89.08         & WDBC      & 93.30 & 95.57 & 53.38 \\
     HMS       &48.07 & 65.90 &  36.56      &  Wine      & 42.10& 49.30 & 34.09 \\
     HDC       & 67.70& 79.90 & 37.80          & Iris      & 18.40 & 21.60 & 14.73 \\
     ION       & 80.40& 86.20 & 31.83         & Seeds     & 30.70 & 37.20 & 19.74 \\
     KVK       &377.60&412.80   & 179.47         & Zoo       &17.50 & 20.30 & 13.59 \\
    \bottomrule
     \end{tabular}
   \end{adjustbox}
\end{table}

In addition, the extracted rule sets from SSL-ART are small for all data sets.  In SSL-ART, the number of labeled prototype nodes indicates the number of rules. In this study, the proportion of labeled samples is only 20\% (for binary classification problems) or 25\% (for multi-class classification problems) of the total training samples.  
As such, the numbers of rules established in SSL-ART (through labeled prototype nodes) are at least three times smaller than those of the unlabeled training samples in binary and multi-class classification problems, respectively.\par

{\color{black}
\subsection{Anomaly detection}
\label{sec:sec:anomaly}
In this section, three artificially generated data sets relevant to anomaly detection are used for performance evaluation, i.e., the Tennessee Eastman process (TEP)~\cite{rieth2018issues}, water distribution (WADI)~\cite{ahmed2017wadi}, and UNSW-NB15~\cite{moustafa2015unsw}. The dimension of these data sets varies from 41 to 121. Specifically, TEP is a model for generating and monitoring faults pertaining to industrial processes in a simulated chemical plant. WADI is collected from a water testbed designed to evaluate water treatment security. UNSW-NB15 captures a wide range of network traffic records, including a mixture of real-world normal and synthesized cyber-attack behaviors.  
Details of these data sets can be found in~\cite{yang2022learning}.\par

WSSL-ART (OtM) and VESSL-ART (OtM) are compared with several models reported in~\cite{yang2022learning}, namely one class SVM (OCSVM), IsolationForest, VAR, AE, DAGMM, OmniAnomaly, GANF, and NiSTAR. Since correctly identifying positive samples is crucial in anomaly detection problems, both F1 measure and accuracy are used for performance evaluation. The experimental results are presented in Table~\ref{Table:anomaly}. Both WESSL-ART (OtM) and VESSL-ART (OTM) yield promising results, as compared with those from other methods in~\cite{yang2022learning}. The superior performance of our proposed models is attributed to their ability to address  overfitting/underfitting issues by avoiding the generation of unnecessary prototype nodes, particularly in the decision boundary regions among different classes. This capability enables our models to produce accurate and robust results, particularly in anomaly detection, as demonstrated in this comparison study.

}

\begin{table}[!h]\color{black}
\vspace{0.4cm}
\centering \caption{\label{Table:anomaly}\color{black}A comparison of accuracy and F1-measure results for anomaly detection with three benchmark data sets reported in~\cite{yang2022learning}. }
    \begin{adjustbox} {width=\columnwidth}
\begin{tabular}{llllllllll}
\toprule
\multirow{2}{*}{Method} &\multicolumn{2}{c}{TEP}   & \multicolumn{2}{c}{WADI} & \multicolumn{2}{c}{UNSW-NB15} \\
\cmidrule{2-7}
               & Accuracy & F1-measure  & Accuracy & F1-measure     & Accuracy & F1-measure   \\
               \midrule
 OCSVM      &  0.6539 & 0.7439 &  0.3495 & 0.9322 & 0.8285  & 0.7305   \\
IsolationForest& 0.6294 & 0.6862 &0.3234 & 0.9300 & 0.8195 & 0.7030  \\
 VAR        &  0.7361 & 0.7780 & 0.3814 & 0.9294 & 0.8189 & 0.7225  \\
 AE         & 0.6582 & 0.7553 & 0.3547 & 0.9180 & 0.8368 & 0.7449   \\
DAGMM      & 0.7030 & 0.7800 & 0.3963 & 0.9150 & 0.8440 & 0.7571  \\
OmniAnomaly& 0.8564 & 0.8833 & 0.5226 & 0.9339 & 0.8487 & 0.7763  \\
GANF       &  0.8808 & 0.9008 & 0.5774 & 0.9448 & 0.8905 & 0.8522 \\
HiSTAR      & \textbf{0.9291} & \textbf{0.9399} & 0.6383 & 0.9541& 0.9122 & 0.8817 \\
            \midrule
VESSL-ART (OtM)& 0.8921  &  0.9113   &  0.6402   &  0.9494   &  \textbf{0.9144}   & 0.8803\\
WESSL-ART (OtM)& 0.9102  &  0.9378   &  \textbf{0.6541}   &  \textbf{0.9572}   &0.9067    & \textbf{0.8824} \\ 
\bottomrule
\end{tabular}
\end{adjustbox}
\end{table}

\vspace*{-0.21cm}
\subsection{A Real-World Case Study}
\label{sec:sec:Motion}
A real-world case study, i.e, human motion detection, has been employed to evaluate the performance of the proposed SSL methods. 
A total of 390 data samples of movement waveform signals have been recorded using 3-axis accelerometers embedded in smartphones~\cite{pourpanah2019animproved}. 
Each subject has been asked to perform two actions: running and walking. 
After pre-processing the waveform signals, nine statistical features have been extracted from each axis, resulting in a total of 27 features (normalized between 0 and 1) for each data sample. The data samples have been divided into learning (80\%) and test (20\%) sets. The learning set has further been partitioned into labeled (25\%) and unlabeled (75\%) samples.\par

To further evaluate the robustness of { WESSL-ART}, two types of noise have been added to the labeled samples.
Firstly, noise has been added to the labels of the data samples. 
A total of 10\% of labeled samples have been randomly selected and switched their class labels. 
Secondly, Gaussian white noise with a signal-to-noise ratio of 10 has been added to the features of the labeled samples. White noise has been added to 10\% of randomly selected learning samples.
In addition, SSL-ART with one-to-one (OtO) mapping is implemented for comparison with the OtM mapping.\par

\subsubsection{Performance evaluation}

Table~\ref{Table:Real-world} presents the results of { WESSL-ART and VESSL-ART} ($M=7$) with both OtM and OtO mapping strategies regarding a number of performance metrics, including correctness, coverage, sensitivity, and specificity.
Overall, the correctness of all models decreases when noise is added to the labeled samples.
The models with $T=3$ produce predictions for more test samples (i.e. a higher degree of coverage) than those with $T=2$.
{  Note that both WESSL-ART and VESSL-ART covers predict a similar number of test samples for each mapping.
In general, WESSL-ART with OTM mapping outperforms other settings. It also can produce balanced sensitivity and specificity rates.
For both WESSL-ART and VESSL-ART, the models with $T=2$ produce better correctness, sensitivity, and specificity than those with $T=3$.
In addition, WESSL-ART and VESSL-ART with OtM mapping perform better than OtO mapping in most instances.}\par

\begin{table*}[!htb]
\centering \caption{\label{Table:Real-world} The experimental results for the human motion detection.}
    \begin{adjustbox} {width=\textwidth}
    \begin{tabular}{l l l c c c c c c c}
    \toprule
      \multirow{2}{*}{Method}& & &  \multicolumn{2}{c}{Noise-free} & \multicolumn{2}{c}{Label (10\%)}& \multicolumn{2}{c}{Feature (10\%)}& \multirow{2}{*}{Average}\\
    \cmidrule{4-9}
                       & &   &   T=2& T=3& T=2& T=3&T=2& T=3\\
                       \cmidrule{1-10}
     \multirow{2}{*}{Coverage} &        
                 \multicolumn{2}{l}{OtO} & 74.61$\pm$ 2.65 & 93.76$\pm$2.12 & 70.07$\pm$2.58 & 84.28$\pm$1.98  & 72.16$\pm$3.26  & 87.59$\pm$2.79 & 80.41$\pm$2.56  \\   
                 
                 & \multicolumn{2}{l}{OtM}   & 78.38$\pm$1.80      & 90.61$\pm$2.10 &72.54$\pm$2.70 & 90.39$\pm$2.67 & 78.38$\pm$ 3.60 & 90.90$\pm$ 2.67 & \hl{83.53$\pm$2.59} \\         
                 
                     \cmidrule{1-10}
    \multirow{3}{*}{VESSL-ART (OtO)} &
                 \multicolumn{2}{l}{Correctness}& 89.45$\pm$ 2.45 &90.54$\pm$2.67 & 82.60$\pm$4.32 & 88.60$\pm$3.61& 88.48$\pm$2.97  & 85.64$\pm$3.70 & 87.55$\pm$3.28  \\
                 & \multicolumn{2}{l}{Sensitivity}& 87.37$\pm$2.83  & 85.34$\pm3.29$ & 76.56$\pm$5.37 & 80.35$\pm$6.36     & 83.57$\pm$ 5.66 & 79.44$\pm$4.68 & 82.10$\pm$4.69  \\
                 & \multicolumn{2}{l}{Specificity}& 94.23$\pm$1.96  & 94.53$\pm$2.60     & 88.56$\pm$4.67 & 96.4$\pm$5.26      & 90.54$\pm$3.98  & 90.31$\pm$6.39 & 92.43$\pm$4.14  \\
                    \cmidrule{1-10}
                    
\multirow{3}{*}{WESSL-ART (OtO)} &
                \multicolumn{2}{l}{Correctness}& 90.64$\pm$ 2.05 &90.67$\pm$3.85 & 84.70$\pm$3.28 & 88.87$\pm$4.18& 86.75$\pm$1.74  & 87.49$\pm$1.36 & 88.18$\pm$3.29  \\
                
                & \multicolumn{2}{l}{Sensitivity}& 89.74$\pm$2.34  & 86.44$\pm3.29$ & 79.63$\pm$2.70 & 83.54$\pm$3.68  & 85.74$\pm$ 6.01 & 83.46$\pm$5.83 & 84.75$\pm$3.97  \\
                
                & \multicolumn{2}{l}{Specificity}& 93.35$\pm$2.62  & 93.34$\pm$3.03 & 90.64$\pm$3.72 & 94.50$\pm$4.61 & 88.21$\pm$4.08  & 89.14$\pm$5.92 & 91.53$\pm$3.99  \\

     \cmidrule{1-10}
    \multirow{3}{*}{VESSL-ART (OtM)} &
                 \multicolumn{2}{l}{Correctness}& 92.18$\pm$2.04 & 89.91$\pm$1.67 &87.21$\pm$3.58 & 89.50$\pm3.31$ & 89.98$\pm$3.67 & 88.15$\pm$3.22 & {89.48$\pm$2.40} \\
                 & \multicolumn{2}{l}{Sensitivity}& 82.93$\pm$2.15      &80.99$\pm$1.18  &81.12$\pm$2.97 & 86.32$\pm$4.23 & 87.07$\pm$4.78 & 84.58$\pm$3.54  & {83.83$\pm$3.14}  \\
                 & \multicolumn{2}{l}{Specificity}& 96.52$\pm$1.94      &94.89$\pm$2.17  &90.60$\pm$3.15 & 92.29$\pm$ 3.67& 91.06$\pm$4.54  & 89.59$\pm$6.32 & {92.49$\pm$3.63}  \\

               \cmidrule{1-10}
    \multirow{3}{*}{WESSL-ART (OtM)} &
                \multicolumn{2}{l}{Correctness}& 94.32$\pm$1.47 & 90.12$\pm$2.73 &88.12$\pm$2.85 & 91.48$\pm2.14$ & 88.81$\pm$2.73 & 90.06$\pm$3.22 & \hl{90.48$\pm$2.52} \\
                & \multicolumn{2}{l}{Sensitivity}& 85.39$\pm$2.54 & 85.30$\pm$2.45  &82.27$\pm$2.97 & 88.21$\pm$4.23 & 85.50$\pm$3.71 & 86.70$\pm$3.54  & \hl{85.56$\pm$3.88}  \\
                & \multicolumn{2}{l}{Specificity}& 95.62$\pm$1.48 & 93.94$\pm$2.74 &93.43$\pm$2.52 & 94.91$\pm$ 1.67& 89.45$\pm$3.65 & 92.19$\pm$2.82 & \hl{93.49$\pm$2.48}  \\

    \bottomrule
     \end{tabular}
    \end{adjustbox}
\end{table*}

\subsubsection{Execution duration and complexity}
{\color{black}Table~\ref{Table:node_real} compares the training durations and the numbers of prototype nodes pertaining to the OtM and OtO mapping schemes on the noise-free (clean) and two noisy (10\% label noise and 10\% feature noise) data sets. 
In the second stage, both mapping schemes generate more prototype nodes corresponding to the training samples that cannot be encoded by the existing prototype nodes generated in the first stage.  
However, OtO generates more prototype nodes than OtM. In addition, the numbers of prototype nodes created by OtM are approximately the same for both noise-free and noisy data sets. 
Comparatively, OtO creates more prototype nodes for handling noisy data. This is mainly attributed to the match-tracking function, which also further increases the execution duration. 
Since the number of unlabeled samples is three times more than labeled ones, the first stage requires a longer execution time.  
Based on Table~\ref{Table:node_real}, it is clear that OtM, as compared with OtO, is a more efficient mapping scheme that generates fewer prototype nodes and consequently requires a shorter execution duration.}\par

\begin{table}[!bt]
\vspace{0.4cm}
\centering \caption{\label{Table:node_real}The  average  number  of  created  prototype  nodes along with execution duration of OtM and OtO mappings.}
    \begin{adjustbox} {width=\columnwidth}
\begin{tabular}{llllllllll}
\toprule
\multirow{3}{*}{Mapping} & &\multicolumn{2}{c}{Clean}   & \multicolumn{2}{c}{Label (10\%)} & \multicolumn{2}{c}{Feature (10\%)} \\
\cmidrule{3-8}
 && Stage 1 & Stage 2  & Stage 1  & Stage 2    & Stage 1   & Stage 2   \\
                          \cmidrule{2-8}
\multirow{2}{*}{OtM} & Nodes      & 55.60 & 74.00  & 57.60 & 72.60 &  59.75 &  73.40\\
                     & Time (ms)  & 95.32 & 40.19 & 97.76 & 37.56 &103.48 & 38.15\\
                     \cmidrule{2-8}
\multirow{2}{*}{OtO} & Nodes      & 55.60 & 79.30  & 57.60 & 83.90 &  59.75 &  87.25 \\
                     & Time (ms)  & 95.32 & 51.76 &97.76 & 55.94 & 103.48 & 60.17\\
\bottomrule
\end{tabular}
\end{adjustbox}
\end{table}

\subsubsection{Coverage vs. correctness}
\begin{figure}
\centering
    \begin{subfigure}[b]{0.24\textwidth}
            \includegraphics[width=\textwidth]{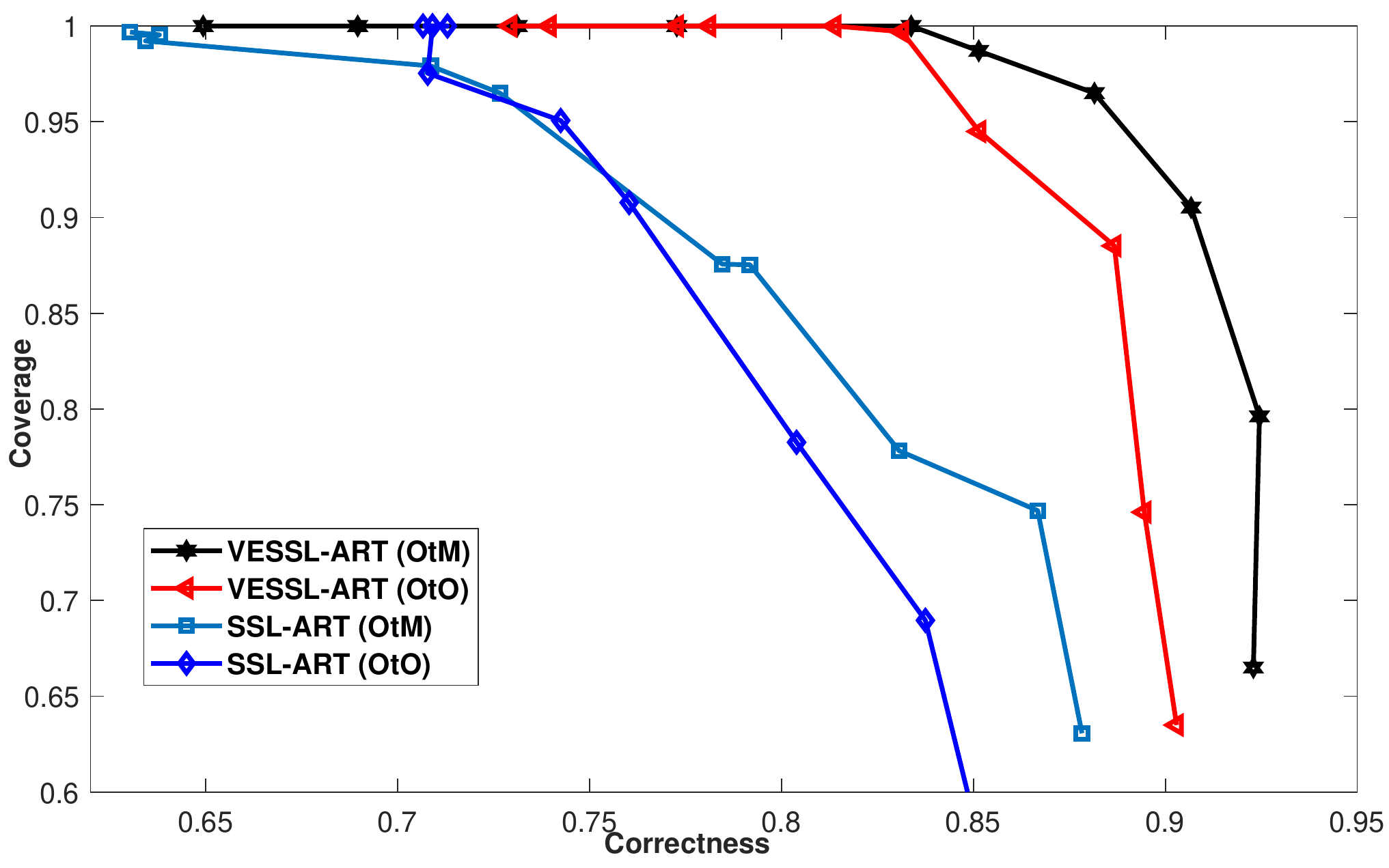}
            \caption{Noise-free (T=2)}
            \label{fig:free2}
    \end{subfigure}%
    \begin{subfigure}[b]{0.24\textwidth}
            \centering
            \includegraphics[width=\textwidth]{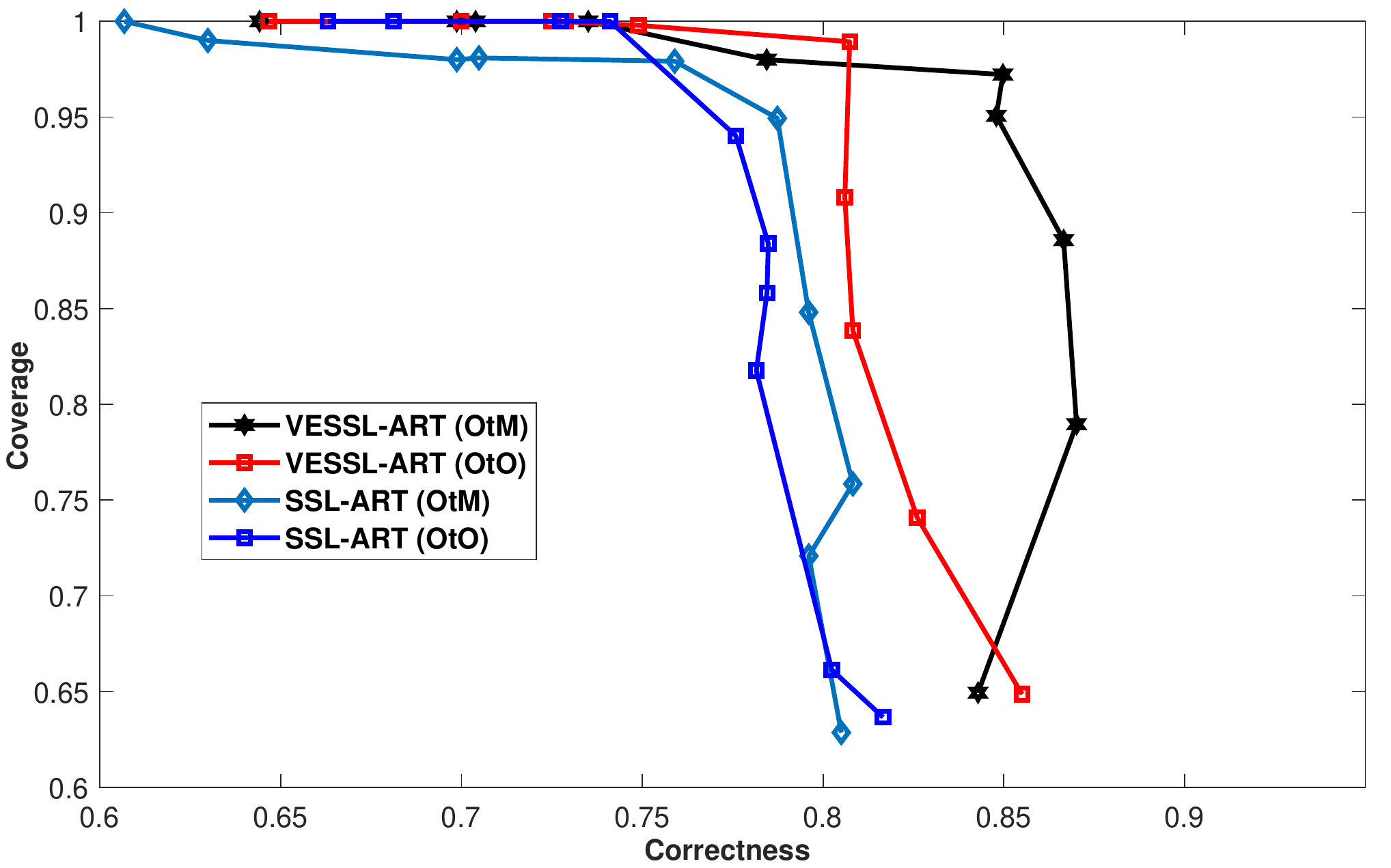}
            \caption{Label 10\% (T=2)}
            \label{fig:lab2}
    \end{subfigure}\\
    \vspace*{0.5cm}
    \begin{subfigure}[b]{0.24\textwidth}
            \centering
            \includegraphics[width=\textwidth]{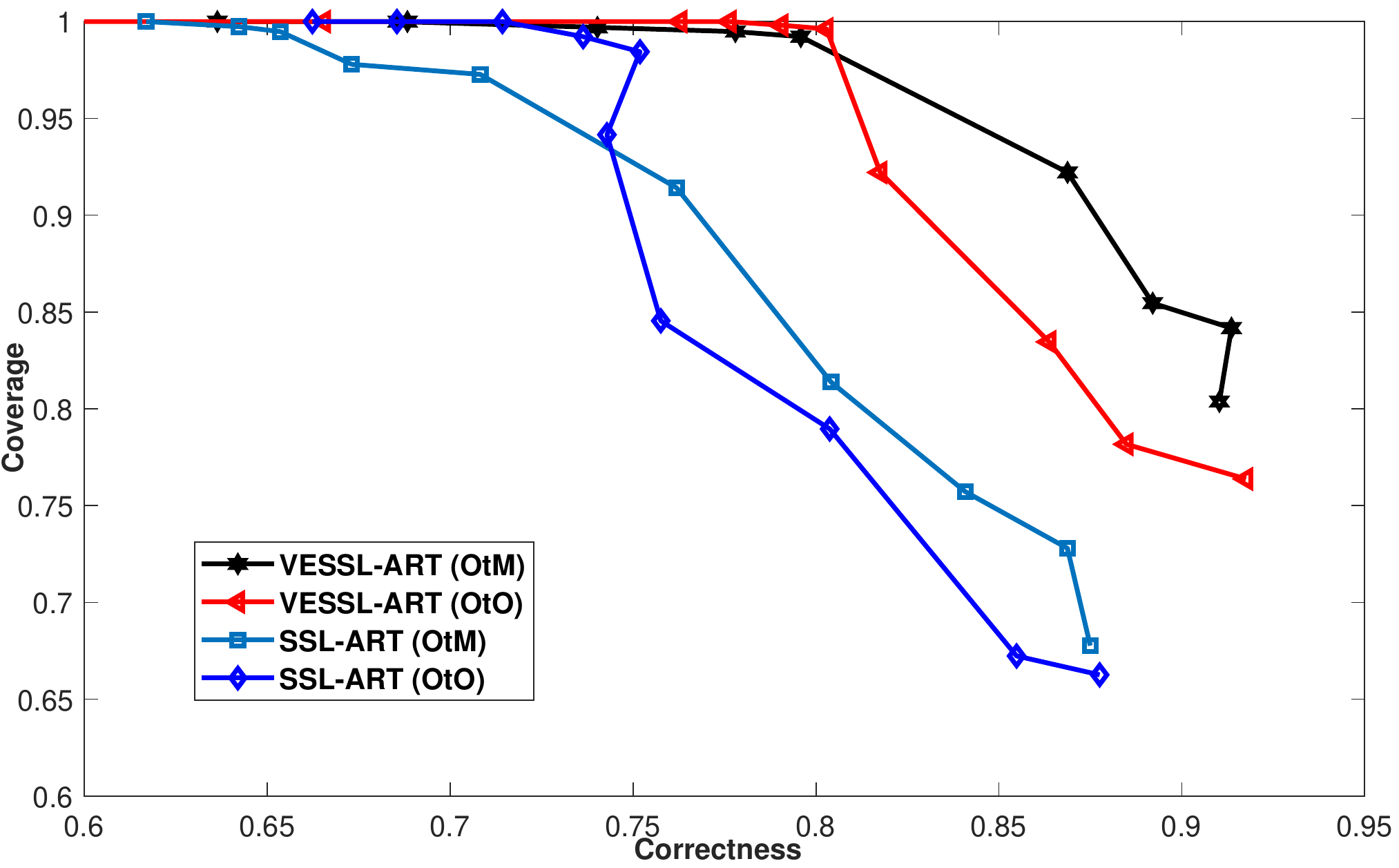}
            \caption{Feature 10\% (T=2)}
            \label{fig:feat2}
    \end{subfigure}
    \begin{subfigure}[b]{0.24\textwidth}
            \centering
            \includegraphics[width=\textwidth]{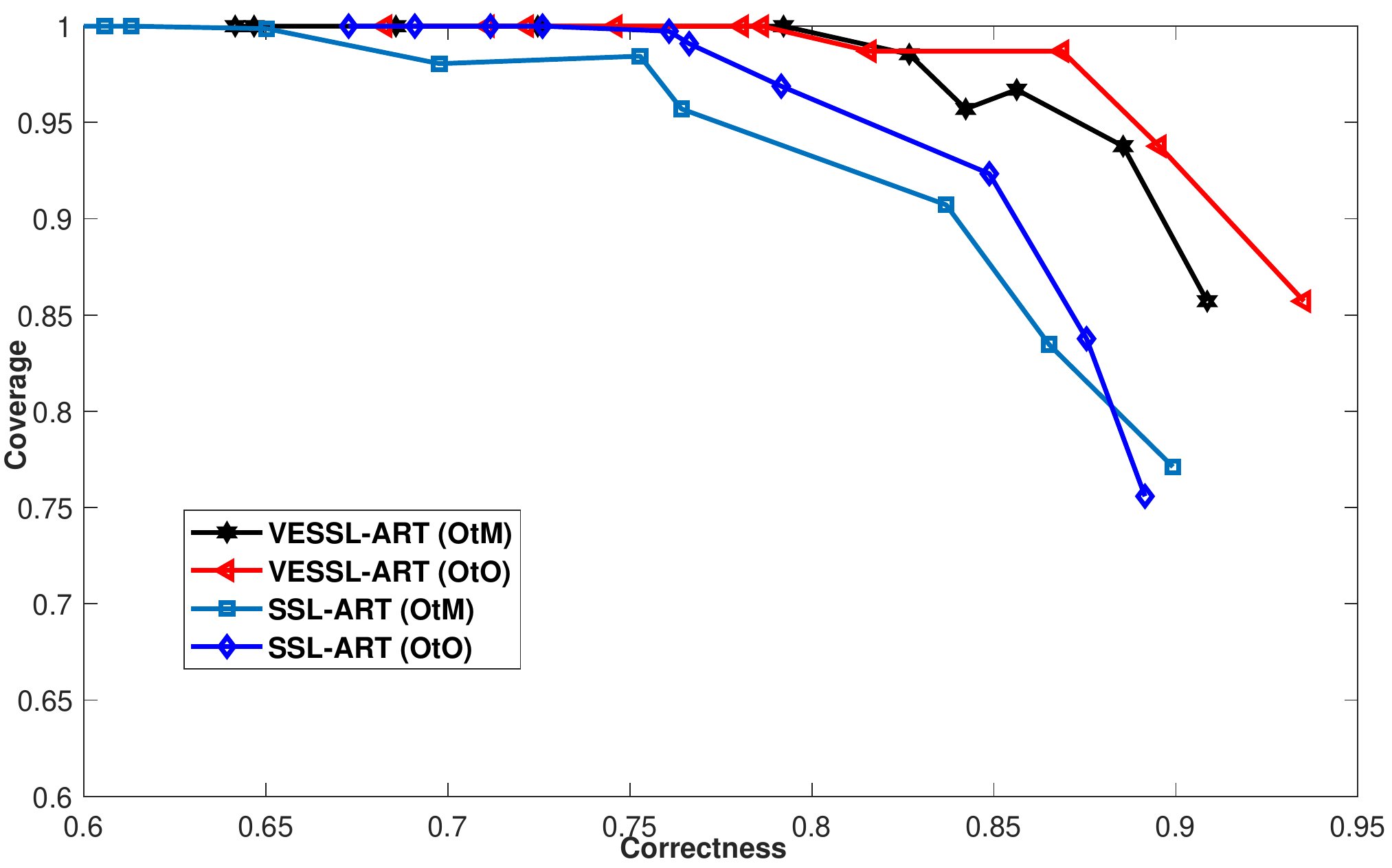}
            \caption{Noise-free (T=3)}
            \label{fig:free3}
    \end{subfigure}\\
    \vspace*{0.5cm}
        \begin{subfigure}[b]{0.24\textwidth}
            \centering
            \includegraphics[width=\textwidth]{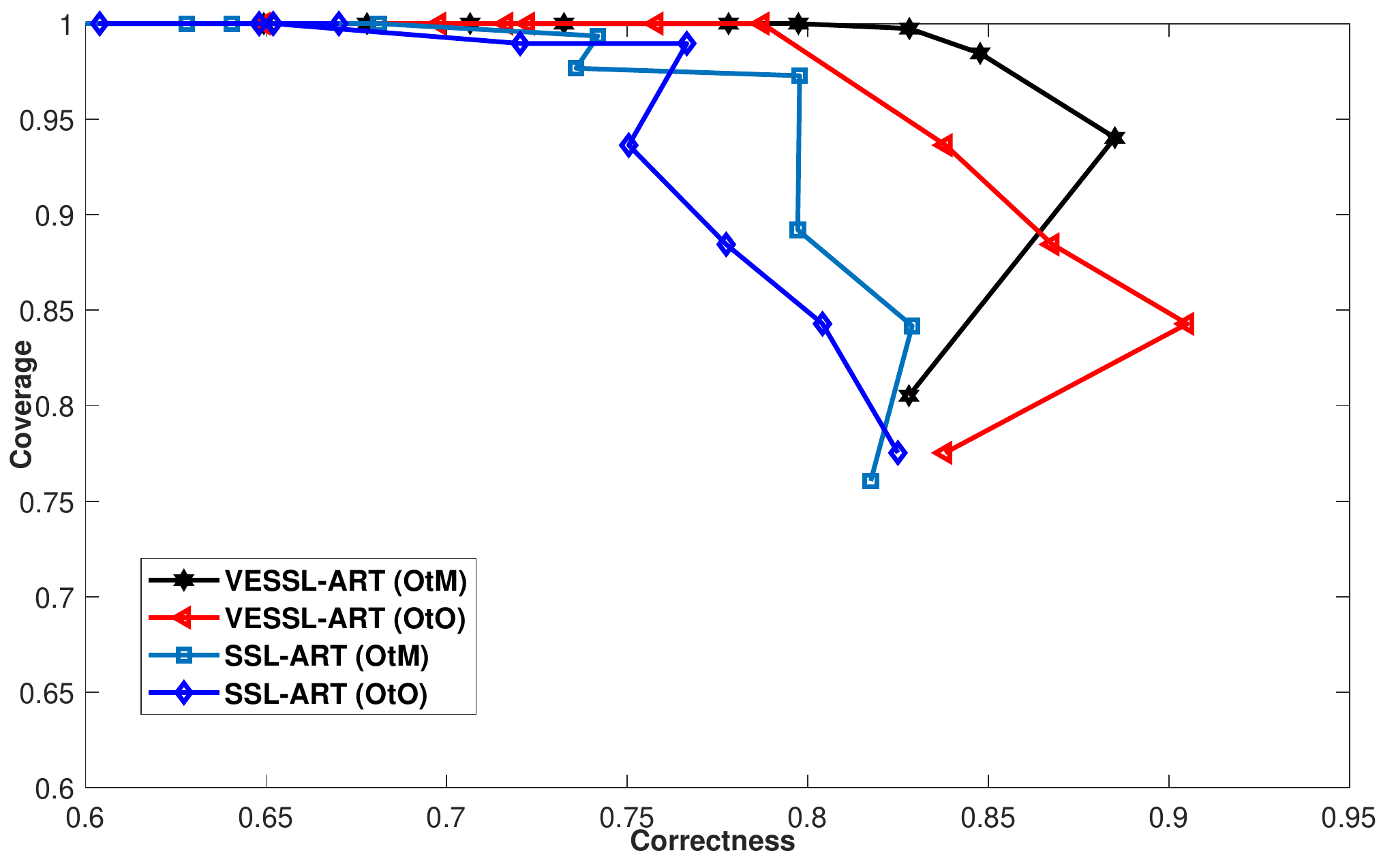}
            \caption{Label 10\% (T=3)}
            \label{fig:lab3}
    \end{subfigure}
        \begin{subfigure}[b]{0.24\textwidth}
            \centering
            \includegraphics[width=\textwidth]{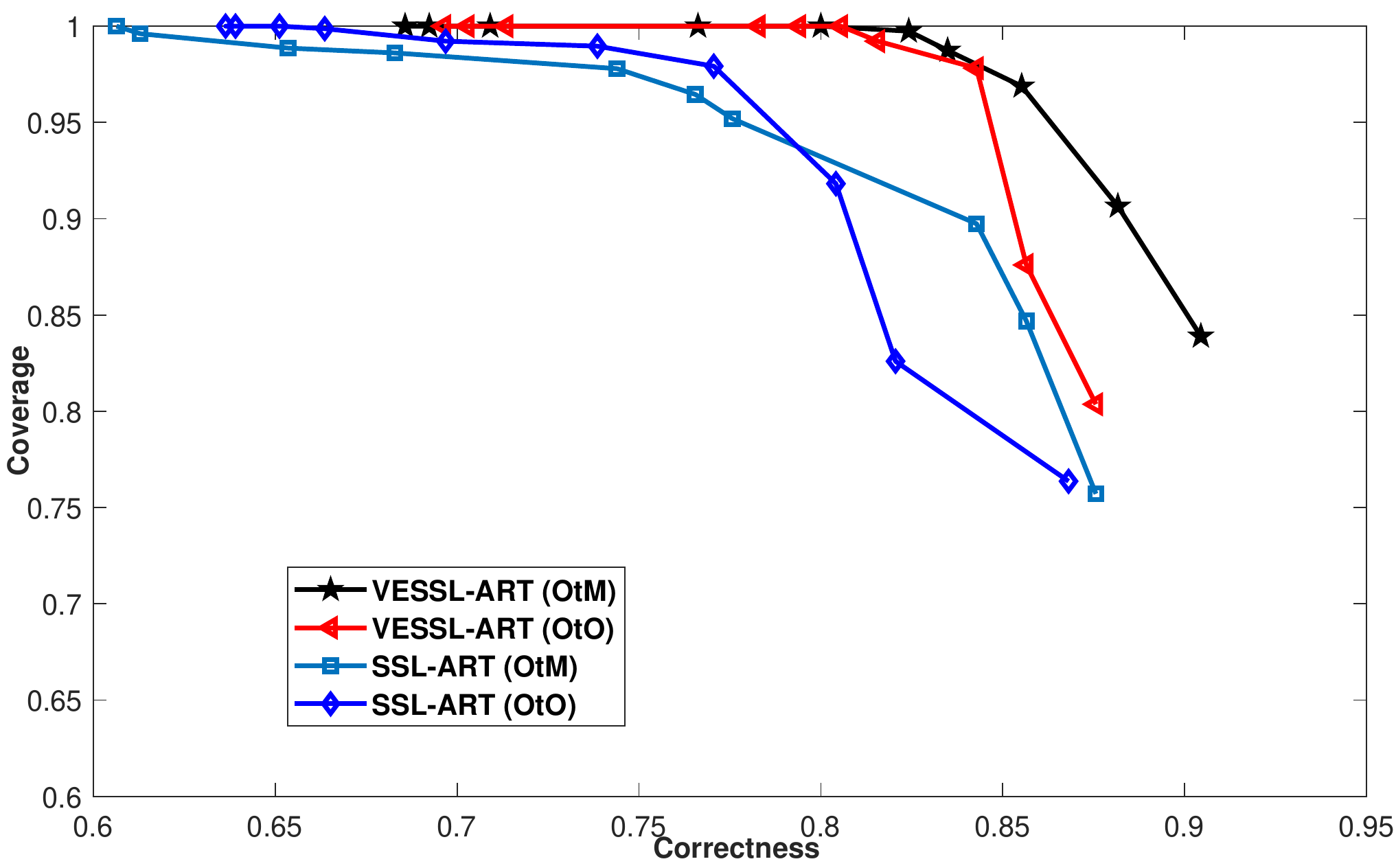}
            \caption{Feature 10\% (T=3)}
            \label{fig:feat3}
    \end{subfigure}
    \vspace*{0.5cm}
    \caption{The trade-off between coverage and correctness for noise-free and noisy (label noise and feature noise at 10\%) data sets.}\label{fig:CorCov}
\end{figure}

Fig.~\ref{fig:CorCov} shows the trade-off between coverage and correctness for both SSL-ART and VESSL-ART models with respect to OtM and OtO mapping strategies using noise-free and noisy data sets.
Overall, VESSL-ART with both mapping strategies are able to provide predictions for the majority of the test samples and yield higher rates of correctness, as compared with those from SSL-ART.
VESSL-ART with OtM mapping outperforms VESSL-ART with OtO mapping in all experiments, except the noise-free data set with $T=3$.
This indicates the usefulness of ESSL-ART with the devised OtM mapping strategy for undertaking noisy (both label noise and feature noise) classification tasks.\par

\vspace*{-0.28cm}
\subsection{Rule Extraction Analysis}
\label{sec:sec:rule}
In this section, the HDC data set is used as an example to demonstrate the rule extraction capability of SSL-ART. 
The input features of the HDC data set include age, sex, CP (chest pain type), TRESTBPS (resting blood pressure), CHOL (serum cholesterol in mg/dl), FBS (fasting blood sugar), RESTECG (resting electrocardiographic results),THALCH (maximum heart rate achieved), XHYPO (1 = yes; 0 = no), OLDPEAK (ST depression induced by exercise relative to rest), SLOPE (slope of the peak exercise ST segment), CA (number of major vessels (0-3) colored by fluoroscopy), THAL (normal=3; fixed defect=6; reversible defect=7).  

\begin{table*}[!ht]
\centering \caption{\label{Table:rule} Four extracted rules for the HDC data set.}
    \begin{adjustbox} {width=\textwidth}
    \begin{tabular}{l c c c c c c c c c c c c c c c}
    \toprule
      No. &  \multicolumn{13}{l}{Features} &Confidence  & Confidence\\
      \cmidrule{2-14}
      & Age & Sex & CP & TRESTBPS & CHOL & FBS & RESTESG & THALCH & XHYPO & OLDPEAK & SLOPE & CA & THAL& class 1 & class 2\\
    \cmidrule{2-16}
    1 & 1 & 1 & 1-2 & 1-2 & 2-3 & 1 & 1 & 2-3 & 1 & 4-5 & 2-3 & 1-2 & 1 & 1.0 & 0.0\\
    2 & 2-3 & 5 & 5 & 2-3 & 2-3 & 1 & 5 & 3 & 1 & 1-2 & 3 & 2 & 5 & 0.0 & 1.0\\
    3 & 2-3 & 1 & 2 & 2 & 1-2 & 5 & 1 & 1-2 & 1 & 1 & 1 & 1-2 & 1 & 0.777 & 0.223\\
    4 & 3-4 & 3 & 5 & 2-3 & 3-4 & 1 & 5 & 2-3 & 1 & 3-4 & 3 & 4 & 1 & 0.428 & 0.572\\
    \bottomrule
     \end{tabular}
    \end{adjustbox}
\end{table*}

The quantization level is set to 5, i.e., ``1=Very Small”, ``2=Small”, ``3=Medium”, ``4=Large”, and ``5=Very Large”, where the coverage of each fuzzy linguistic variable can be determined in consultation with domain experts.  
Four extracted rules are shown in Table~\ref{Table:rule}.  
As an example, the first rule can be interpreted as follows:\\

\textbf{If} \textit{Age} is ``Very Small", AND

~~~\textit{Sex} is ``Very Small", AND

~~~\textit{CP} is from ``Small" to ``Very Small", AND

~~~\textit{TRESTBPS} is from ``Very Small" to ``Small", AND

~~~\textit{CHOL} is from ``Small" to ``Medium", AND

~~~\textit{FBS} is ``Very Small", AND 

~~~\textit{RESTECG} is ``Very Small", AND

~~~\textit{THALCH} is from ``Small" to ``Medium", AND

~~~\textit{XHYPO} is ``Very Small", AND

~~~\textit{OLDPEAK} is from ``Large" to ``Very Large", AND

~~~\textit{SLOPE} is from ``Small" to ``Medium", AND

~~~\textit{CA} is from ``Very Small" to ``Small", AND

~~~\textit{THAL} is ``Very Small"

\textbf{Then} Positive for HDC with confidence estimate=1.0. 
\vspace*{0.3cm}

Note that the performance measure of SSL-ART (i.e. classification accuracy) is not affected by the fuzzy representation during rule extraction.  
This is because rule extraction is performed after the learning phase of SSL-ART (where the best accuracy rate has been achieved).  
The aim of rule extraction is to provide justification/explanation for users to better understand the predicted outcome from SSL-ART, in order to mitigate the black-box nature of most neural network-based learning models.\par

Specifically, the accuracy score of SSL-ART is largely determined by number of prototype nodes created, which is governed by the vigilance setting.  
A smaller vigilance parameter results in fewer prototype nodes with larger hyperbox sizes (i.e. coarser classification granularity), while a larger vigilance parameter leads to more prototype nodes with smaller hyperbox sizes (i.e. finer classification granularity.  
However, a larger number of prototype nodes (i.e. a more complex network) does not guarantee a better classification accuracy, but may cause over-fitting. 
Similarly, under-fitting may occur when a smaller number of prototype nodes (i.e. a simpler network) is formed, compromising classification accuracy.\par

\vspace*{-0.3cm}
\section{Conclusion}
\label{sec:Conclusion}
In this paper, semi-supervised ART-based learning models, namely SSL-ART and WESSL-ART, have been proposed.  
The unsupervised fuzzy ART network is used as the underlying model to generate a number of prototype nodes using unlabeled samples. 
Then, leveraging the supervised fuzzy ARTMAP structure, labeled samples are employed to label the created prototype nodes.  
During both unsupervised and supervised learning processes, SSL-ART performs one-pass learning through both labeled and unlabeled samples, while new prototype nodes can be added incrementally whenever necessary. An OtM mapping scheme has been formulated to associate each prototype node with more than one class label, in order to reduce the number of redundant prototype nodes as well as to reduce the effects of noisy samples.
{ 
In addition, a weighted voting strategy has been introduced to form an ensemble model of WESSL-ART.  
Each ensemble member, i.e. SSL-ART, assigns a different weight to each class.  
The final predicted class is selected based on a weighting voting scheme.  
The aim is to mitigate the effects of training data sequences on the SSL-ART members and to improve the overall performance of WESSL-ART.  
Through a series of empirical studies, the outcome indicates that SSL-ART and WESSL-ART are able to improve their performance using unlabeled samples and produce comparable results as compared with those from other methods reported in the literature.  
Useful If-Then rules have also been extracted to provide justification of the prediction to the users.\par

For further research, we will focus on improving the OtM mapping scheme that can lead to the loss of information.  SSL-ART labels each prototype node to belong to only one target class that has accumulated the highest number of associative links with respect to all target classes, while ignoring the remaining target classes that may contain correct, clean, and/or noise-free samples.  
In particular, this phenomenon can happen in the decision boundary regions where there exist data samples from different classes. 
We will investigate the use of reinforcement learning or probability-based method to address this problem, in order to enhance the robustness of the proposed ART-based SSL models.

}

\ifCLASSOPTIONcaptionsoff
  \newpage
\fi



%

\vspace*{-0.1cm}
\bibliographystyle{IEEEtran}

\bibliography{mybib}

\end{document}